\documentclass[10pt,twocolumn,letterpaper]{article}

\usepackage{cvpr}
\usepackage{times}
\usepackage{epsfig}
\usepackage{graphicx}
\usepackage{amsmath}
\usepackage{amssymb}
\usepackage{stackengine}

\cvprfinalcopy 

\def\cvprPaperID{****} 

\ifcvprfinal\fi
\begin{document}

\title{Inception-v4, Inception-ResNet and\\the Impact of Residual Connections on Learning}

\author{Christian Szegedy\\
Google Inc.\\
1600 Amphitheatre Pkwy, Mountain View, CA\\
{\tt\small szegedy@google.com}
\and
Sergey Ioffe\\
{\tt\small sioffe@google.com}
\and
Vincent Vanhoucke\\
{\tt\small vanhoucke@google.com}
\and
Alex Alemi\\
{\tt\small alemi@google.com}
}

\maketitle

\begin{abstract}
Very deep convolutional networks have been central to the largest advances in image
recognition performance in recent years. One example is the Inception
architecture that has been shown to achieve very good performance at relatively
low computational cost. Recently, the introduction of residual connections
in conjunction with a more traditional architecture has yielded state-of-the-art
performance in the 2015 ILSVRC challenge; its performance was similar
to the latest generation Inception-v3 network. This raises the question of whether
there are any benefit in combining the Inception architecture with residual
connections.
Here we give clear empirical evidence that training with residual connections
accelerates the training of Inception networks significantly. There is also
some evidence of residual Inception networks outperforming similarly
expensive Inception networks without residual connections by a thin margin.
We also present several new streamlined architectures for both residual and
non-residual Inception networks. These variations improve the single-frame
recognition performance on the ILSVRC 2012 classification task significantly.
We further demonstrate how proper activation scaling stabilizes the training of
very wide residual Inception networks. With an ensemble of three residual and
one Inception-v4, we achieve 3.08\% top-5 error on the test set of
the ImageNet classification (CLS) challenge.

\end{abstract}

\section{Introduction}

Since the 2012 ImageNet competition~\cite{russakovsky2014imagenet}
winning entry by Krizhevsky et al~\cite{krizhevsky2012imagenet},
their network ``AlexNet'' has been successfully applied to a larger variety of
computer vision tasks, for example to object-detection~\cite{girshick2014rcnn},
segmentation~\cite{long2015fully}, human pose estimation~\cite{toshev2014deeppose},
video classification~\cite{karpathy2014large}, object
tracking~\cite{wang2013learning}, and superresolution~\cite{dong2014learning}.
These examples are but a few of all the applications to which deep
convolutional networks have been very successfully applied ever since.

In this work we study the combination of the two most recent ideas:
Residual connections introduced by He et al. in ~\cite{he2015deep} and the latest
revised version of the Inception architecture~\cite{szegedy2015rethinking}.
In~\cite{he2015deep}, it is argued that residual connections are of inherent
importance for training very deep architectures. Since Inception networks
tend to be very deep, it is natural to replace the
filter concatenation stage of the Inception architecture with residual connections. This
would allow Inception to reap all the benefits of the residual approach
while retaining its computational efficiency.

Besides a straightforward integration, we have also studied whether
Inception itself can be made more efficient by making it deeper and wider.
For that purpose, we designed a new version named Inception-v4
which has a more uniform simplified architecture and more inception modules
than Inception-v3. Historically, Inception-v3 had inherited a lot of the
baggage of the earlier incarnations. The technical constraints chiefly came from
the need for partitioning the model for distributed training using
DistBelief~\cite{dean2012large}.
Now, after migrating our training setup to TensorFlow~\cite{tensorflow2015-whitepaper}
these constraints have been lifted, which allowed us to simplify the architecture
significantly. The details of that simplified architecture are described in Section \ref{arch}.

In this report, we will compare the two pure Inception variants,
Inception-v3 and v4, with similarly expensive hybrid Inception-ResNet
versions. Admittedly, those models were picked in a somewhat ad hoc manner
with the main constraint being that the parameters and computational
complexity of the models should be somewhat similar to the cost
of the non-residual models. In fact we have tested bigger and wider
Inception-ResNet variants and they performed very similarly on the
ImageNet classification challenge ~\cite{russakovsky2014imagenet}
dataset.

The last experiment reported here is an evaluation of an ensemble of
all the best performing models  presented here. As it was
apparent that both Inception-v4 and Inception-ResNet-v2 performed
similarly well, exceeding state-of-the art single frame performance
on the ImageNet validation dataset, we wanted to see how a combination
of those pushes the state of the art on this well studied dataset.
Surprisingly, we found that gains on the single-frame performance do not
translate into similarly large gains on ensembled performance. Nonetheless,
it still allows us to report 3.1\% top-5 error on the validation set with
four models ensembled setting a new state of the art, to our best
knowledge.

In the last section, we study some of the classification failures and
conclude that the ensemble still has not reached the label noise of
the annotations on this dataset and there is still room for improvement
for the predictions.

\section{Related Work}

Convolutional networks have become popular in large scale image recognition
tasks after Krizhevsky et al.~\cite{krizhevsky2012imagenet}. Some of the next important
milestones were Network-in-network~\cite{lin2013network} by Lin et al.,
VGGNet~\cite{simonyan2014very} by Simonyan et al. and GoogLeNet
(Inception-v1)~\cite{szegedy2015going} by Szegedy et al.

Residual connection were introduced by He et al. in~\cite{he2015deep} in
which they give convincing theoretical and practical evidence for the
advantages of utilizing additive merging of signals both for image recognition, and especially for object detection.
The authors argue that residual connections are inherently necessary for training
very deep convolutional models. Our findings do not seem to support this
view, at least for image recognition. However it might require more
measurement points with deeper architectures to understand the true extent
of beneficial aspects offered by residual connections.
In the experimental section we demonstrate that it is not very difficult to
train competitive very deep networks without utilizing residual connections.
However the use of residual connections seems to improve the training speed
greatly, which is alone a great argument for their use.
\begin{figure}
\centering
\includegraphics[width=0.5\linewidth]{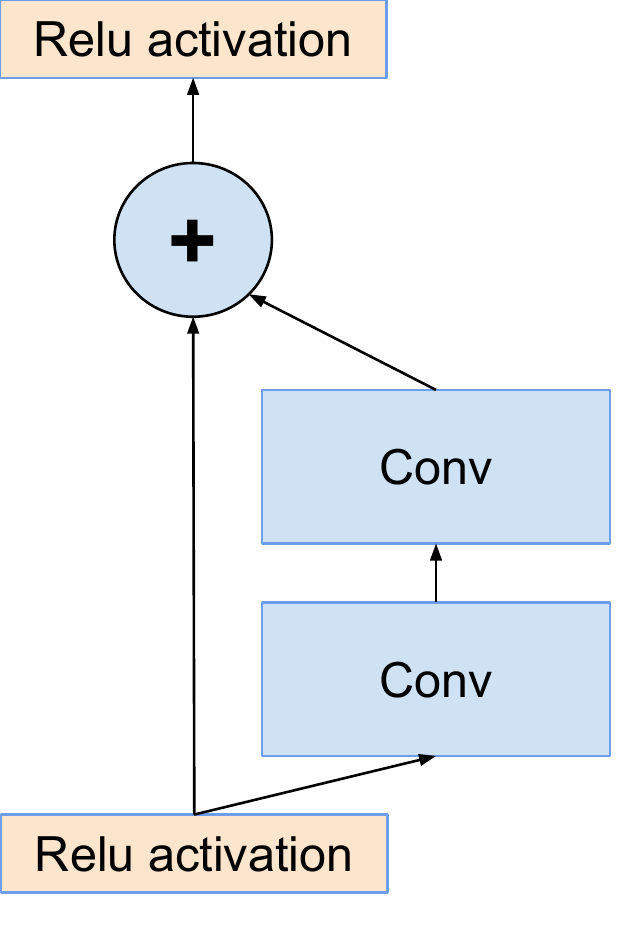}
\caption{Residual connections as introduced in He et al.~\cite{he2015deep}.}
\label{fig:resnetsimple}
\end{figure}
\begin{figure}
\centering
\includegraphics[width=0.5\linewidth]{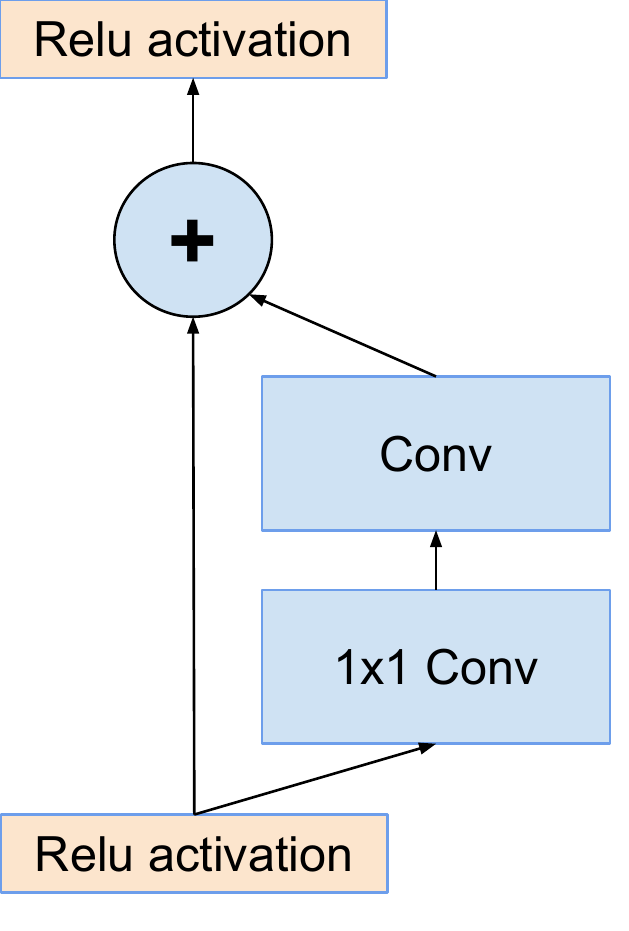}
\caption{Optimized version of ResNet connections by~\cite{he2015deep} to
  shield computation.}
\label{fig:resnetoptimized}
\end{figure}
The Inception deep convolutional architecture was introduced
in~\cite{szegedy2015going} and was called GoogLeNet or Inception-v1 in our
exposition.
Later the Inception architecture was refined in various ways,
first by the introduction of batch normalization
~\cite{ioffe2015batch} (Inception-v2) by Ioffe et al.
Later the architecture was improved by additional factorization ideas in the
third iteration~\cite{szegedy2015rethinking} which will be referred to as
Inception-v3 in this report.

\section{Architectural Choices}
\label{arch}

\subsection{Pure Inception blocks}

\begin{figure}
\centering
\includegraphics[width=0.7\linewidth]{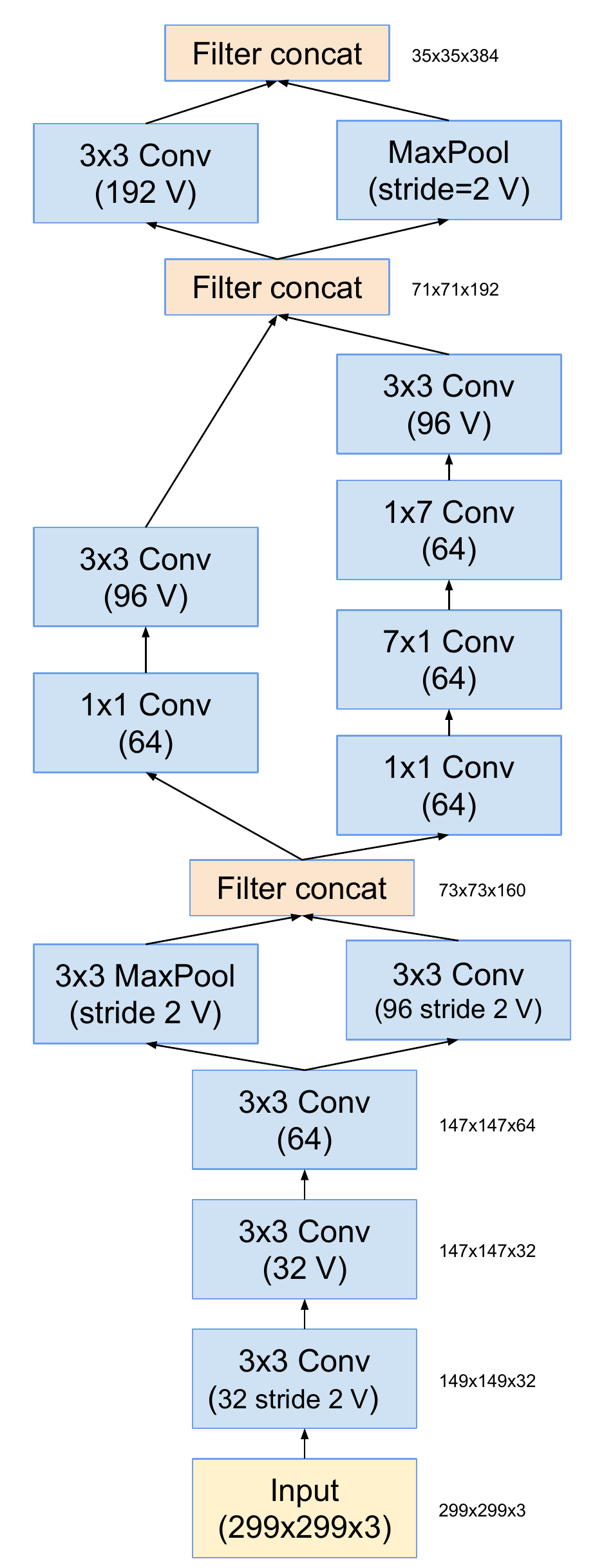}
\caption{The schema for stem of the pure Inception-v4 and
  Inception-ResNet-v2 networks. This is the input part of those
networks. Cf. Figures~\ref{fig:inceptionv4} and~\ref{fig:resnetsmallschema} }
\label{fig:inceptionv4stem}
\end{figure}

\begin{figure}
\centering
\includegraphics[width=\linewidth]{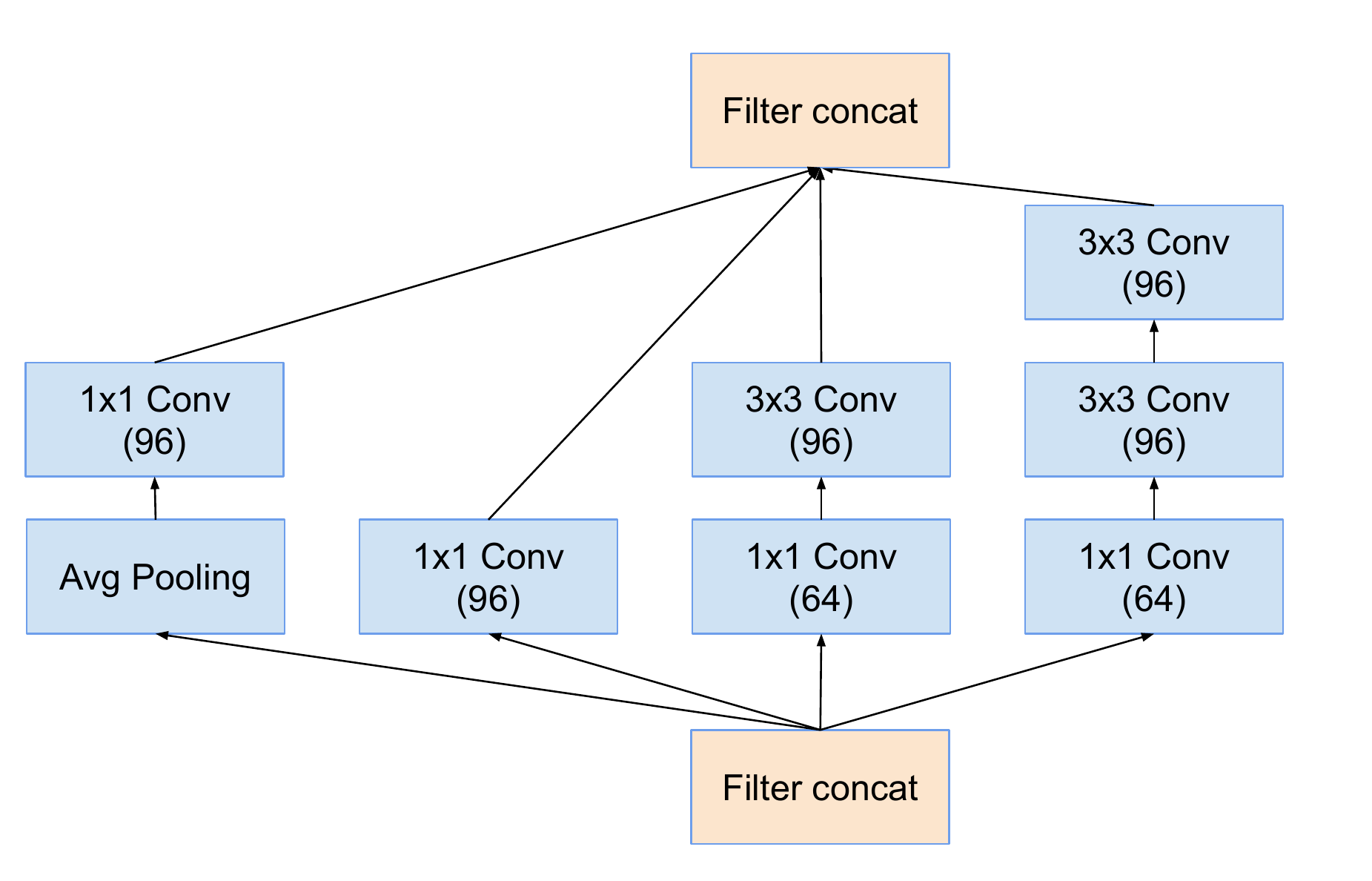}
\caption{The schema for $35\times 35$ grid modules of the pure Inception-v4
 network. This is the Inception-A block of Figure~\ref{fig:inceptionv4}. }
\label{fig:wide35x35module}
\end{figure}

\begin{figure}
\centering
\includegraphics[width=\linewidth]{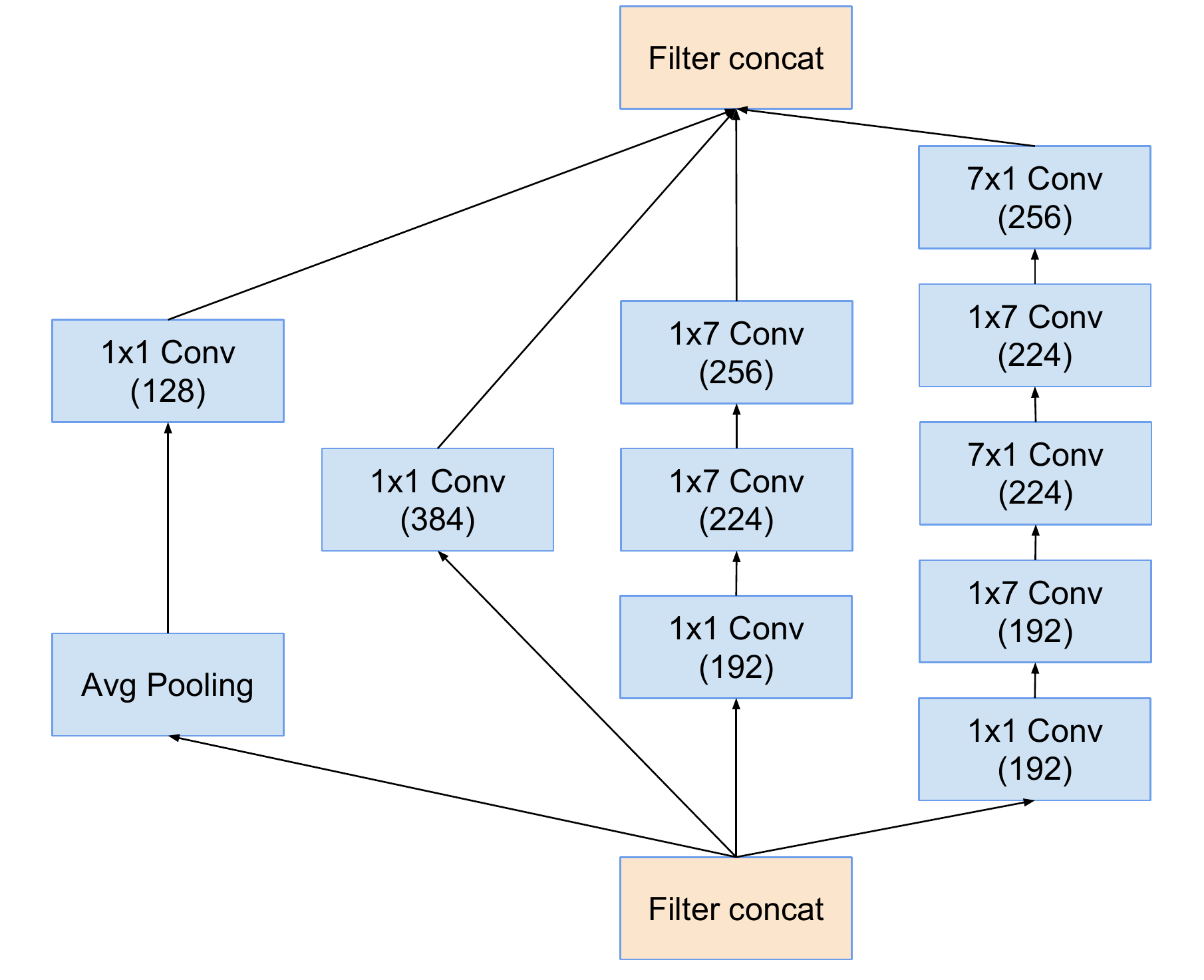}
\caption{The schema for $17\times 17$ grid modules of the pure Inception-v4
 network. This is the Inception-B block of Figure~\ref{fig:inceptionv4}.}
\label{fig:wide17x17module}
\end{figure}

\begin{figure}
\centering
\includegraphics[width=\linewidth]{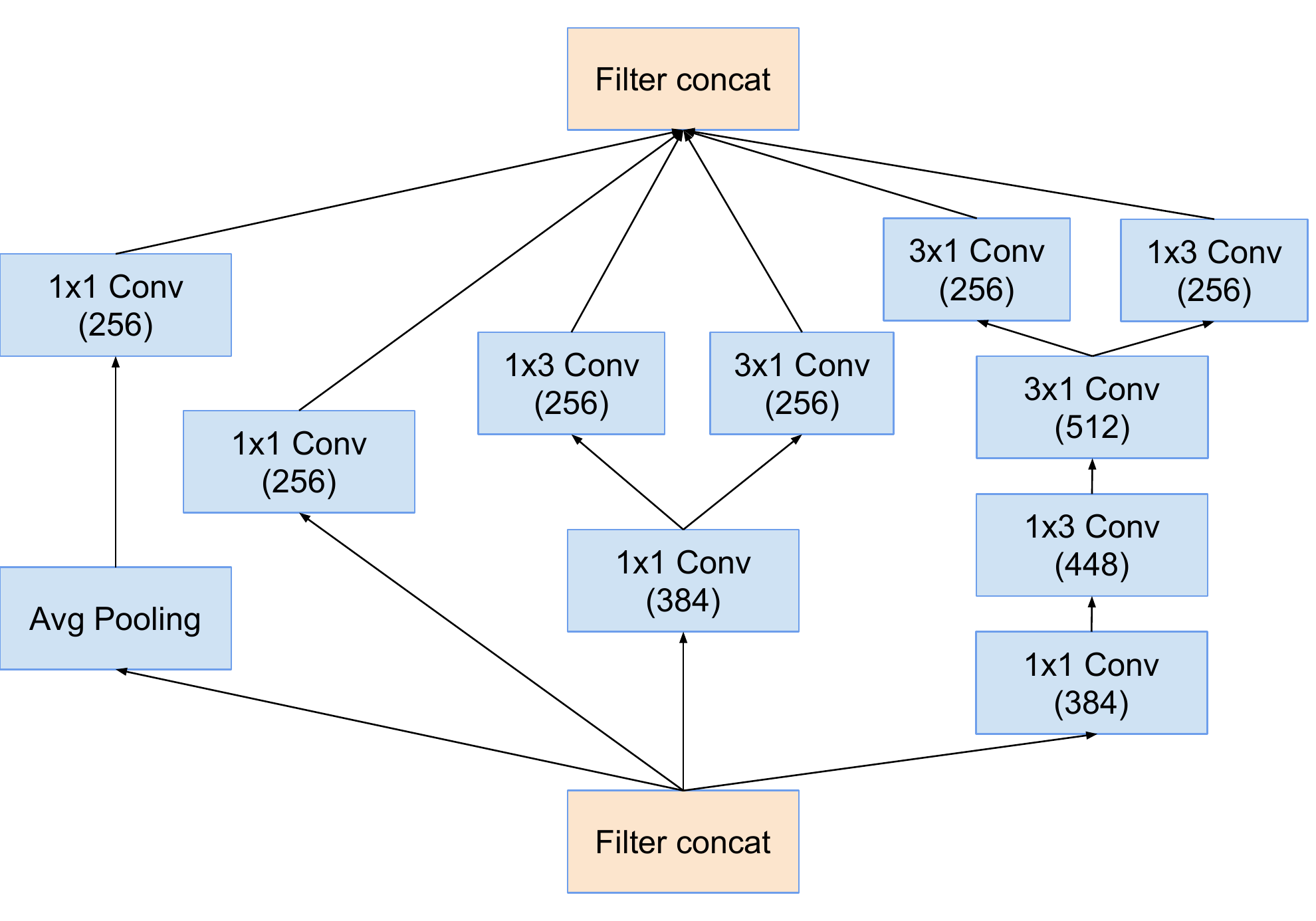}
\caption{The schema for $8\times 8$ grid modules of the pure Inception-v4
 network. This is the Inception-C block of Figure~\ref{fig:inceptionv4}.}
\label{fig:wide8x8module}
\end{figure}

\begin{figure}
\centering
\includegraphics[width=\linewidth]{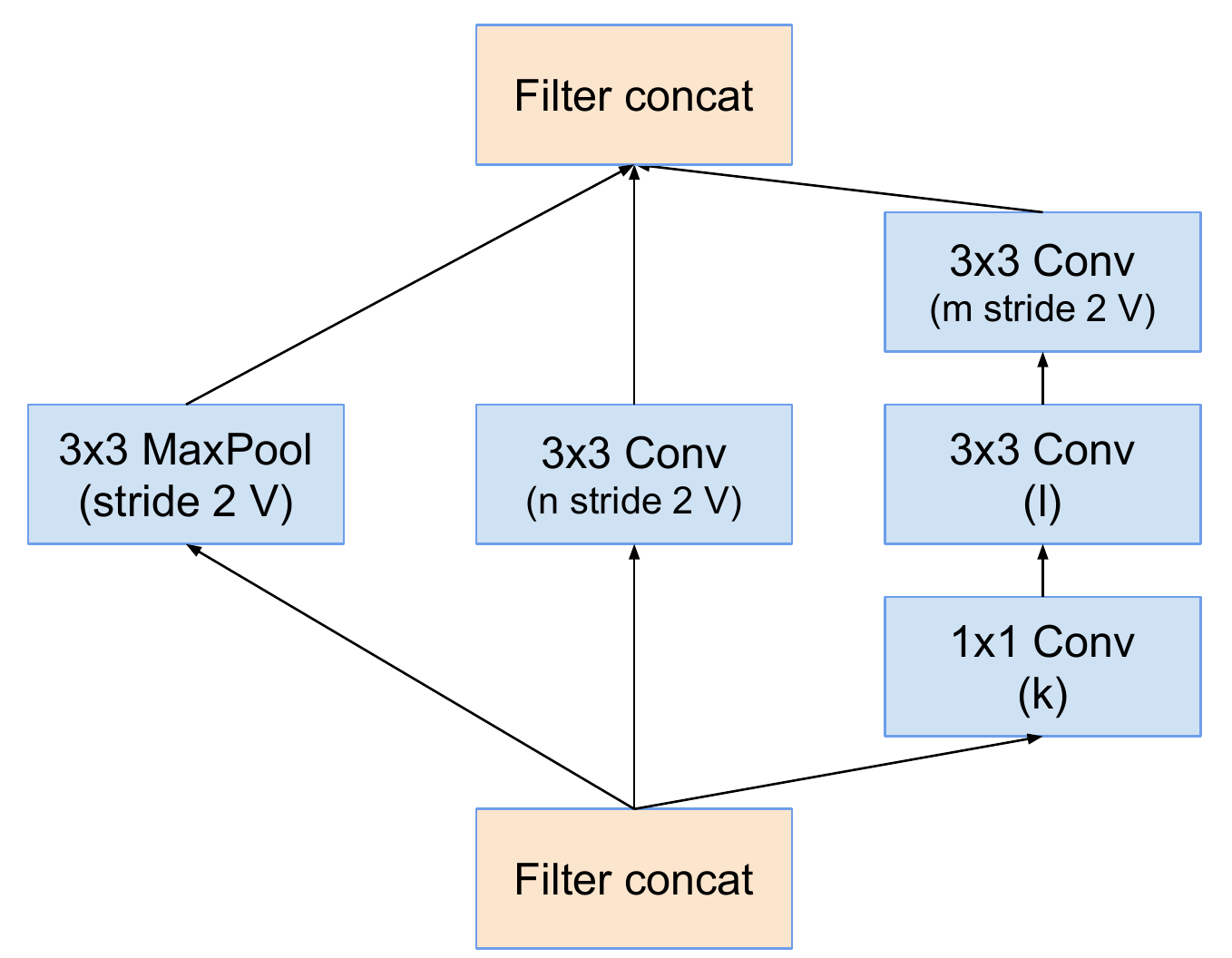}
\caption{The schema for $35\times 35$ to $17\times 17$ reduction module.
  Different variants of this blocks (with various number of filters) are
  used in Figure~\ref{fig:inceptionv4}, and \ref{fig:resnetsmallschema}
  in each of the new Inception(-v4, -ResNet-v1, -ResNet-v2) variants
  presented in this paper. The $k$, $l$, $m$, $n$ numbers represent
  filter bank sizes which can be looked up in Table~\ref{reductionto17params}.
}
\label{fig:reductionto17}
\end{figure}

\begin{figure}
\centering
\includegraphics[width=\linewidth]{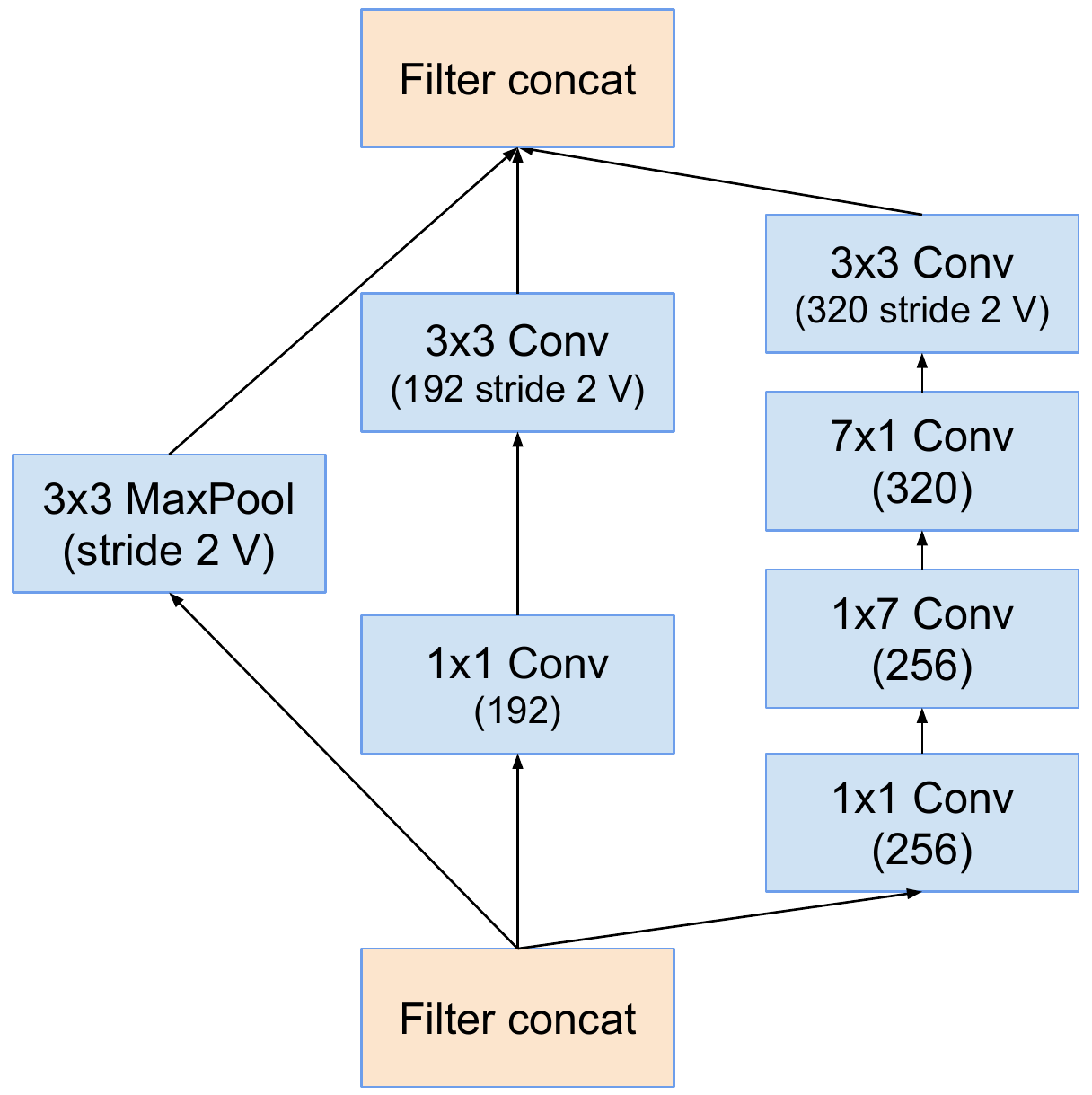}
\caption{The schema for $17\times 17$ to $8\times 8$ grid-reduction module.
  This is the reduction module used by the pure Inception-v4 network in
  Figure~\ref{fig:inceptionv4}.
}
\label{fig:reductionto8}
\end{figure}

\begin{figure}
\centering
\includegraphics[width=0.5\linewidth]{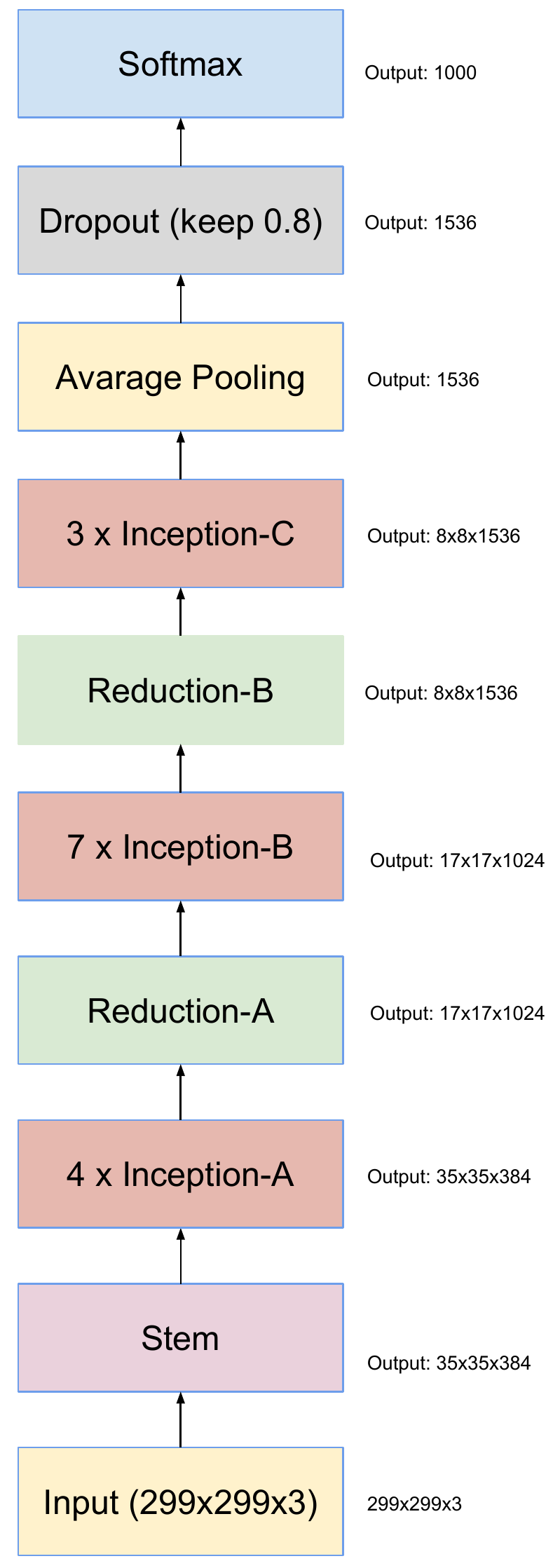}
\caption{The overall schema of the Inception-v4 network. For the
  detailed modules, please refer to Figures~\ref{fig:inceptionv4stem},
  \ref{fig:wide35x35module}, \ref{fig:wide17x17module}, \ref{fig:wide8x8module},
  \ref{fig:reductionto17} and \ref{fig:reductionto8} for the detailed structure
  of the various components.
}
\label{fig:inceptionv4}
\end{figure}

Our older Inception models used to be trained in a partitioned manner,
where each replica was partitioned into a multiple sub-networks in order to be able to
fit the whole model in memory. However, the Inception architecture is highly
tunable, meaning that there are a lot of possible changes to the number of filters
in the various layers that do not affect the quality of the fully trained
network. In order to optimize the training speed, we used to tune the layer sizes
carefully in order to balance the computation between the various model
sub-networks.
In contrast, with the introduction of TensorFlow our most recent models
can be trained without partitioning the replicas. This is enabled in part by recent
optimizations of memory used by backpropagation, achieved by carefully considering
what tensors are needed for gradient computation and structuring the computation
to reduce the number of such tensors. Historically, we have been
relatively conservative about changing the architectural choices and restricted
our experiments to varying isolated network components while keeping the
rest of the network stable. Not simplifying earlier choices
resulted in networks that looked more complicated that they needed to be.
In our newer experiments, for Inception-v4 we decided to shed this unnecessary
baggage and made uniform choices for the Inception blocks for each grid size.
Plase refer to Figure~\ref{fig:inceptionv4} for the large scale
structure of the Inception-v4 network and Figures~\ref{fig:inceptionv4stem},
\ref{fig:wide35x35module}, \ref{fig:wide17x17module}, \ref{fig:wide8x8module},
\ref{fig:reductionto17} and \ref{fig:reductionto8} for the detailed structure
of its components. All the convolutions not marked with ``V'' in the figures
are same-padded meaning that their output grid matches the size of their input.
Convolutions marked with ``V'' are valid padded, meaning that input patch of
each unit is fully contained in the previous layer and the grid size of the
output activation map is reduced accordingly.

\subsection{Residual Inception Blocks}
For the residual versions of the Inception networks, we use cheaper Inception
blocks than the original Inception. Each Inception block is followed by
filter-expansion layer ($1\times 1$ convolution without activation) which is
used for scaling up the dimensionality of the filter bank before the addition
to match the depth of the input. This is needed to compensate for the dimensionality
reduction induced by the Inception block.

We tried several versions of the residual version of Inception. Only two
of them are detailed here. The first one ``Inception-ResNet-v1''
roughly the computational cost of Inception-v3, while ``Inception-ResNet-v2''
matches the raw cost of the newly introduced Inception-v4 network. See
Figure~\ref{fig:resnetsmallschema} for the large scale structure of both
varianets. (However, the step time of Inception-v4 proved to be significantly
slower in practice, probably due to the larger number of layers.)

Another small technical difference between our residual and non-residual
Inception variants is that in the case of Inception-ResNet,
we used batch-normalization only on top of the traditional layers, but not
on top of the summations. It is reasonable to expect that a thorough use
of batch-normalization should be advantageous, but we wanted to keep
each model replica trainable on a single GPU. It turned out that the
memory footprint of layers with large activation size was consuming
disproportionate amount of GPU-memory. By omitting the batch-normalization
on top of those layers, we were able to increase the overall number of
Inception blocks substantially. We hope that with better utilization of
computing resources, making this trade-off will become unecessary.

\begin{figure}
\centering
\includegraphics[width=\linewidth]{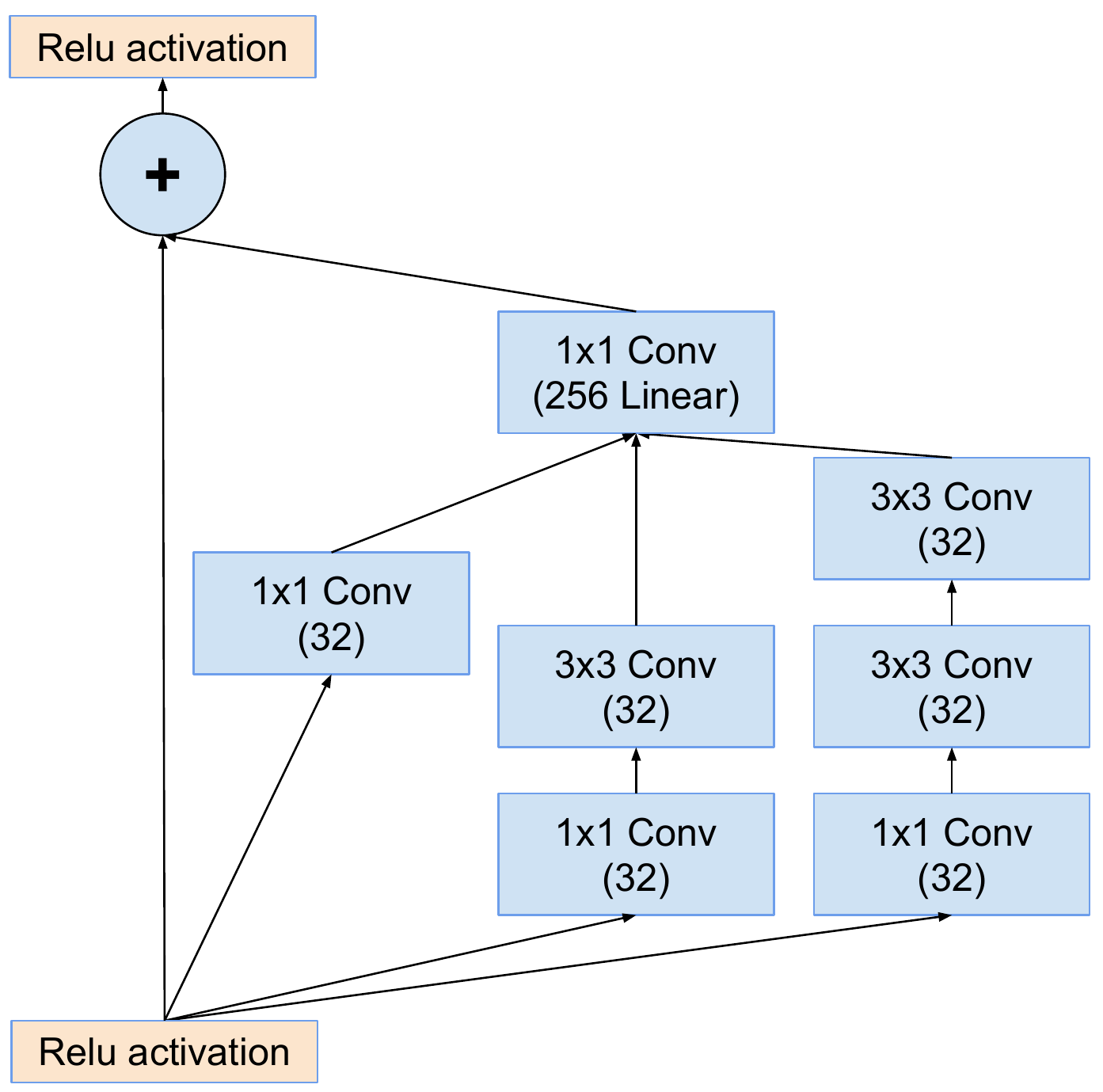}
\caption{The schema for $35\times 35$ grid (Inception-ResNet-A) module of Inception-ResNet-v1
 network.}
\label{fig:resnetsmall35x35module}
\end{figure}
\begin{figure}
\centering
\includegraphics[width=\linewidth]{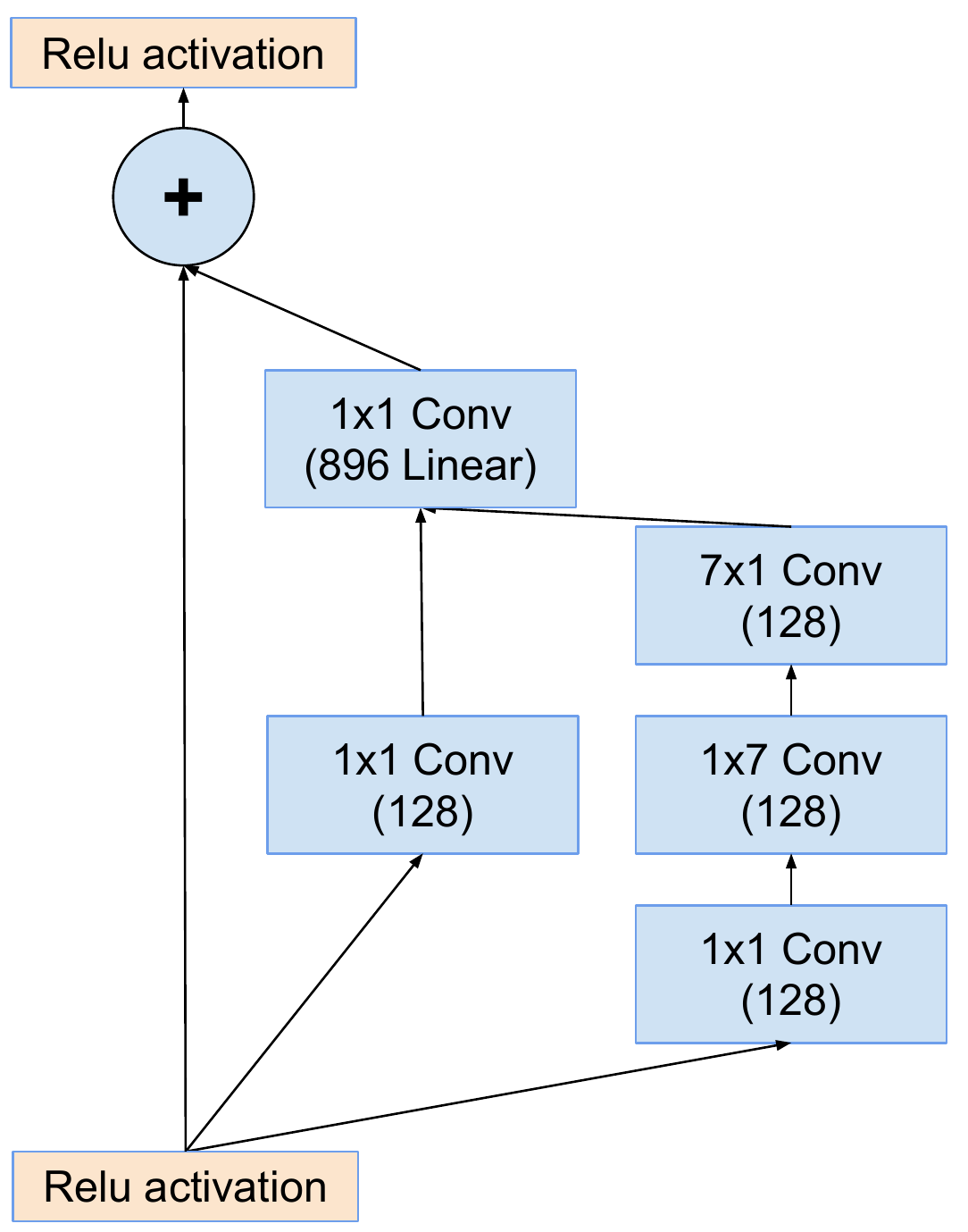}
\caption{The schema for $17\times 17$ grid (Inception-ResNet-B)  module of Inception-ResNet-v1
 network.}
\label{fig:resnetsmall17x17module}
\end{figure}
\begin{figure}
\centering
\includegraphics[width=\linewidth]{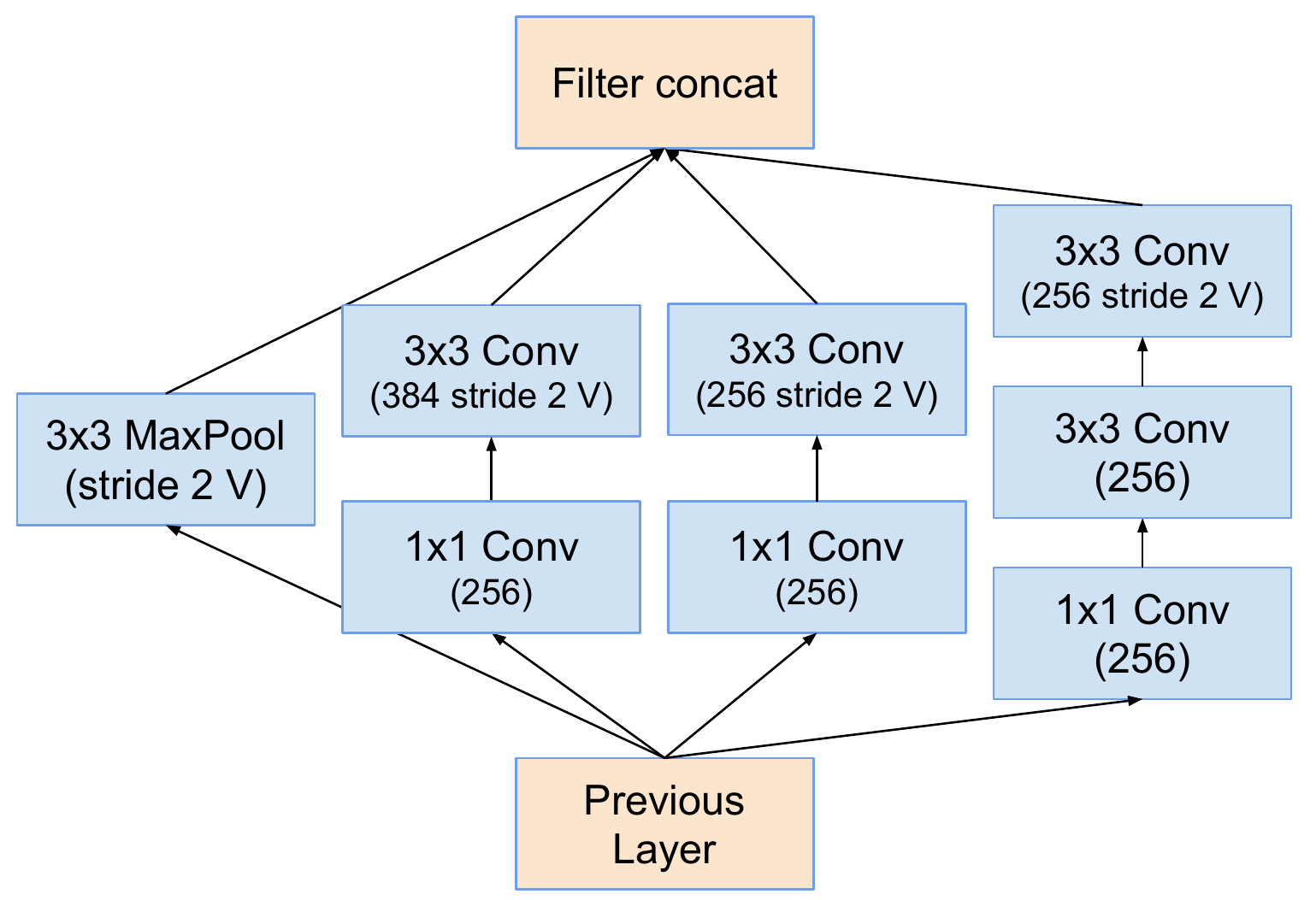}
\caption{``Reduction-B'' $17\times 17$ to $8\times 8$ grid-reduction module.
  This module used by the smaller Inception-ResNet-v1 network
  in Figure~\ref{fig:resnetsmallschema}.
}
\label{fig:reductionto8resnet}
\end{figure}
\begin{figure}
\centering
\includegraphics[width=\linewidth]{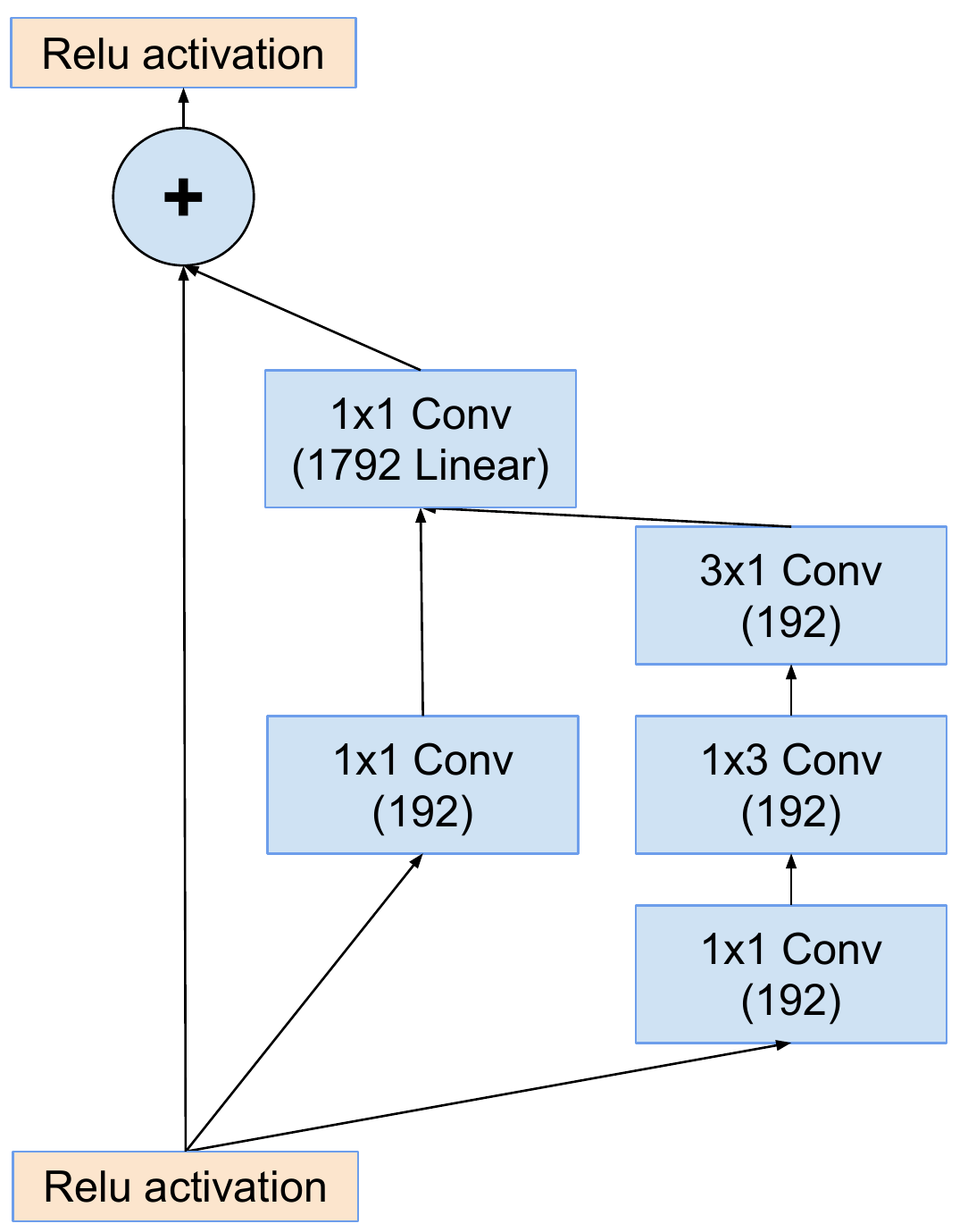}
\caption{The schema for $8\times 8$ grid (Inception-ResNet-C) module of Inception-ResNet-v1
 network.}
\label{fig:resnetsmall8x8module}
\end{figure}
\begin{figure}
\centering
\includegraphics[width=0.5\linewidth]{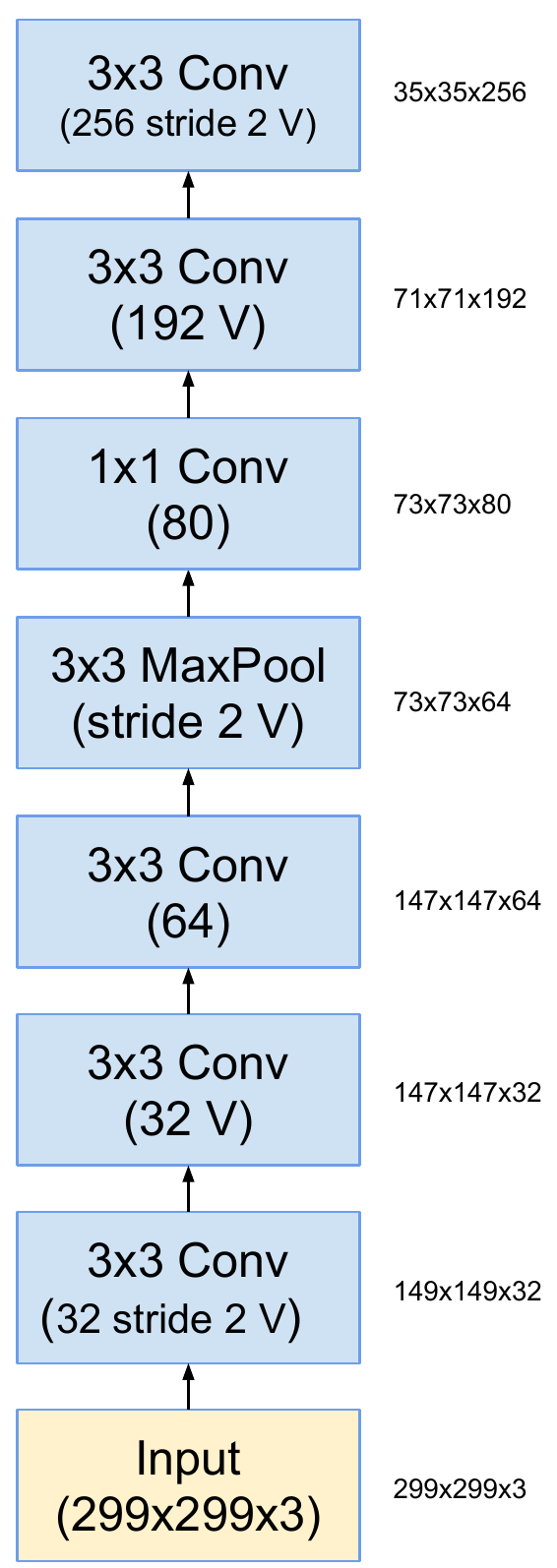}
\caption{The stem of the Inception-ResNet-v1 network.}
\label{fig:resnetsmallstem}
\end{figure}
\begin{figure}
\centering
\includegraphics[width=0.5\linewidth]{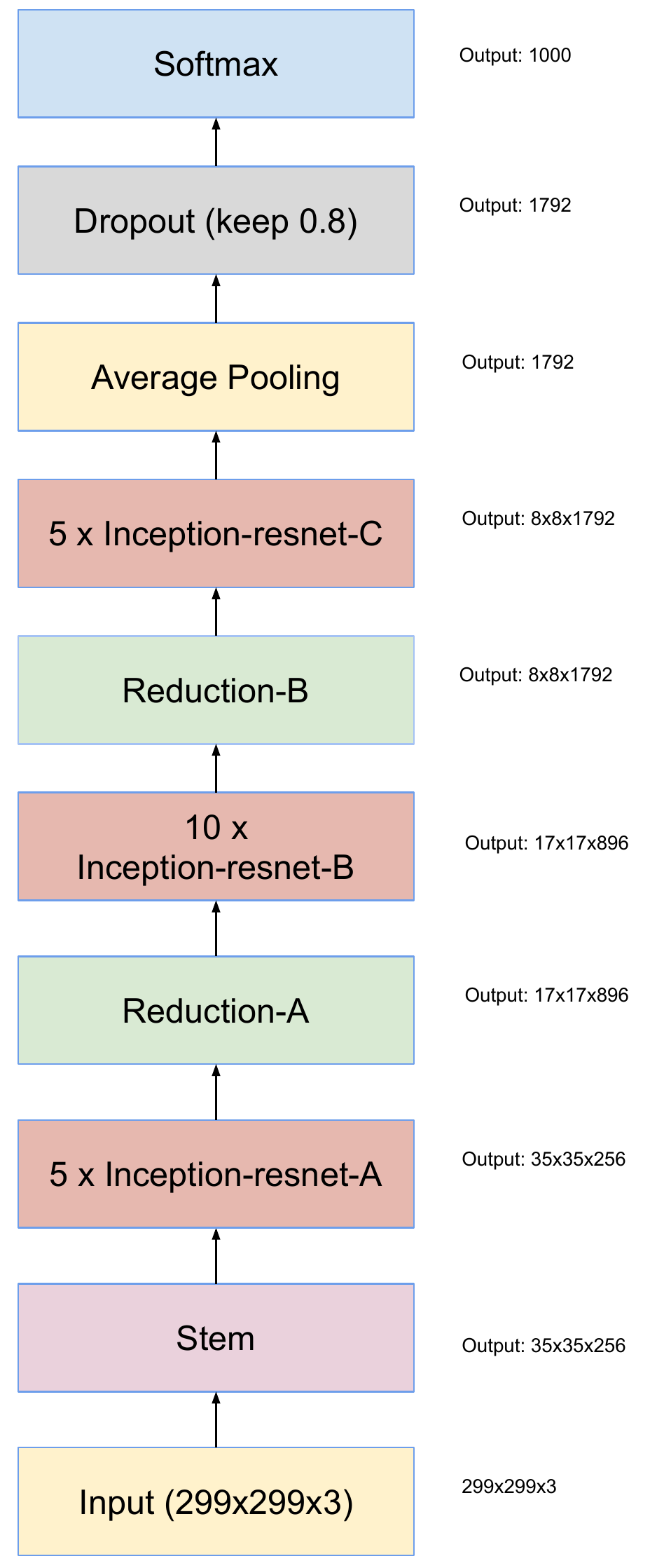}
\caption{Schema for Inception-ResNet-v1 and Inception-ResNet-v2 networks.
  This schema applies to both networks but the underlying components differ.
  Inception-ResNet-v1 uses the blocks as described in Figures~\ref{fig:resnetsmallstem},
  \ref{fig:resnetsmall35x35module}, \ref{fig:reductionto17}, \ref{fig:resnetsmall17x17module},
  \ref{fig:reductionto8resnet} and \ref{fig:resnetsmall8x8module}.
  Inception-ResNet-v2 uses the blocks as described in Figures~\ref{fig:inceptionv4stem},
  \ref{fig:resnetwide35x35module}, \ref{fig:reductionto17},\ref{fig:resnetwide17x17module},
  \ref{fig:reductionto8resnetwide} and \ref{fig:resnetwide8x8module}.
  The output sizes in the diagram refer to the activation vector tensor shapes of
  Inception-ResNet-v1.
}
\label{fig:resnetsmallschema}
\end{figure}
\clearpage
\begin{figure}
\centering
\includegraphics[width=\linewidth]{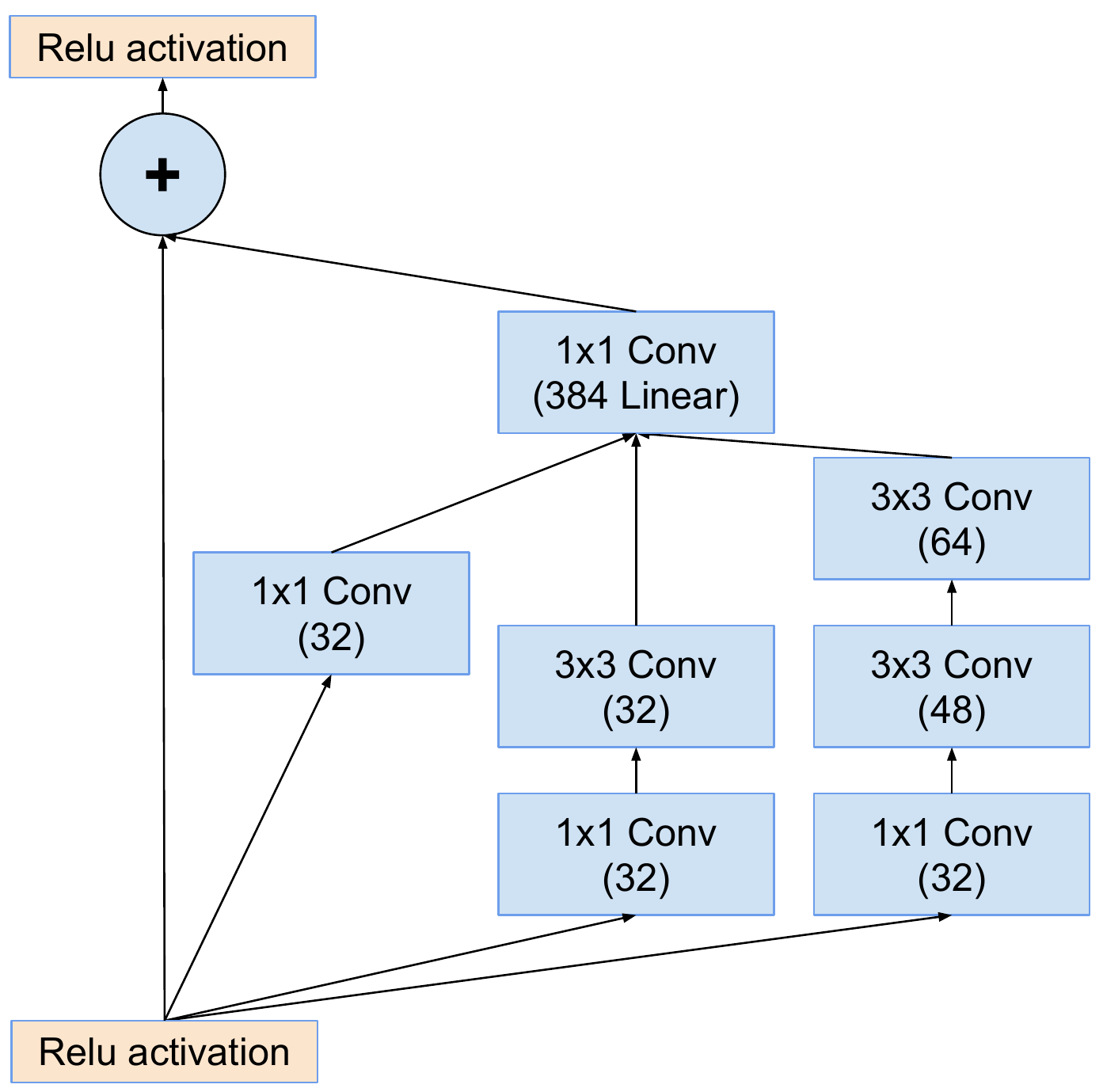}\caption{The schema for
  $35\times 35$ grid (Inception-ResNet-A) module of the Inception-ResNet-v2 network.}
\label{fig:resnetwide35x35module}
\end{figure}
\begin{figure}
\centering
\includegraphics[width=\linewidth]{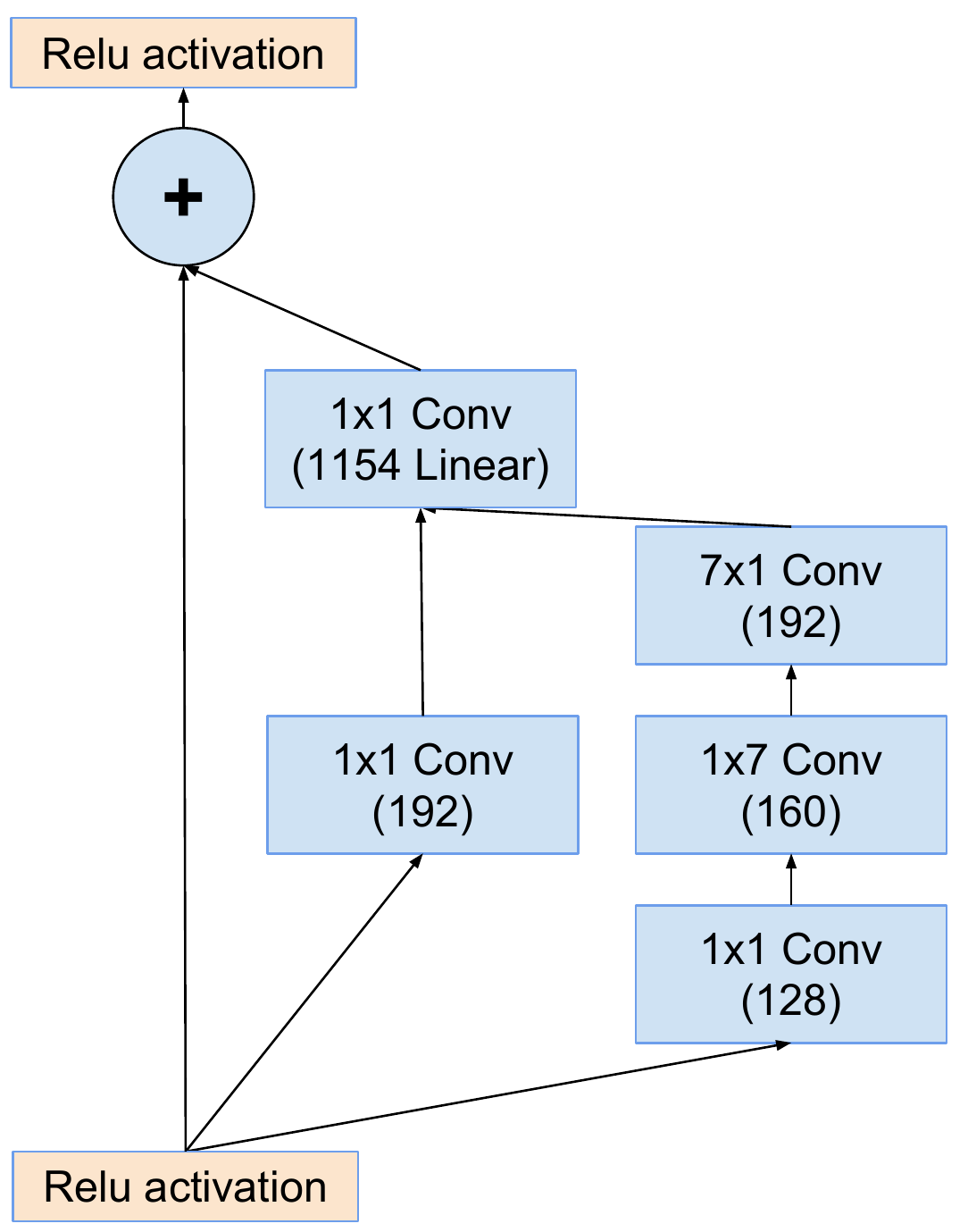}
\caption{The schema for $17\times 17$ grid (Inception-ResNet-B) module of the
  Inception-ResNet-v2  network.}
\label{fig:resnetwide17x17module}
\end{figure}
\begin{figure}
\centering
\includegraphics[width=\linewidth]{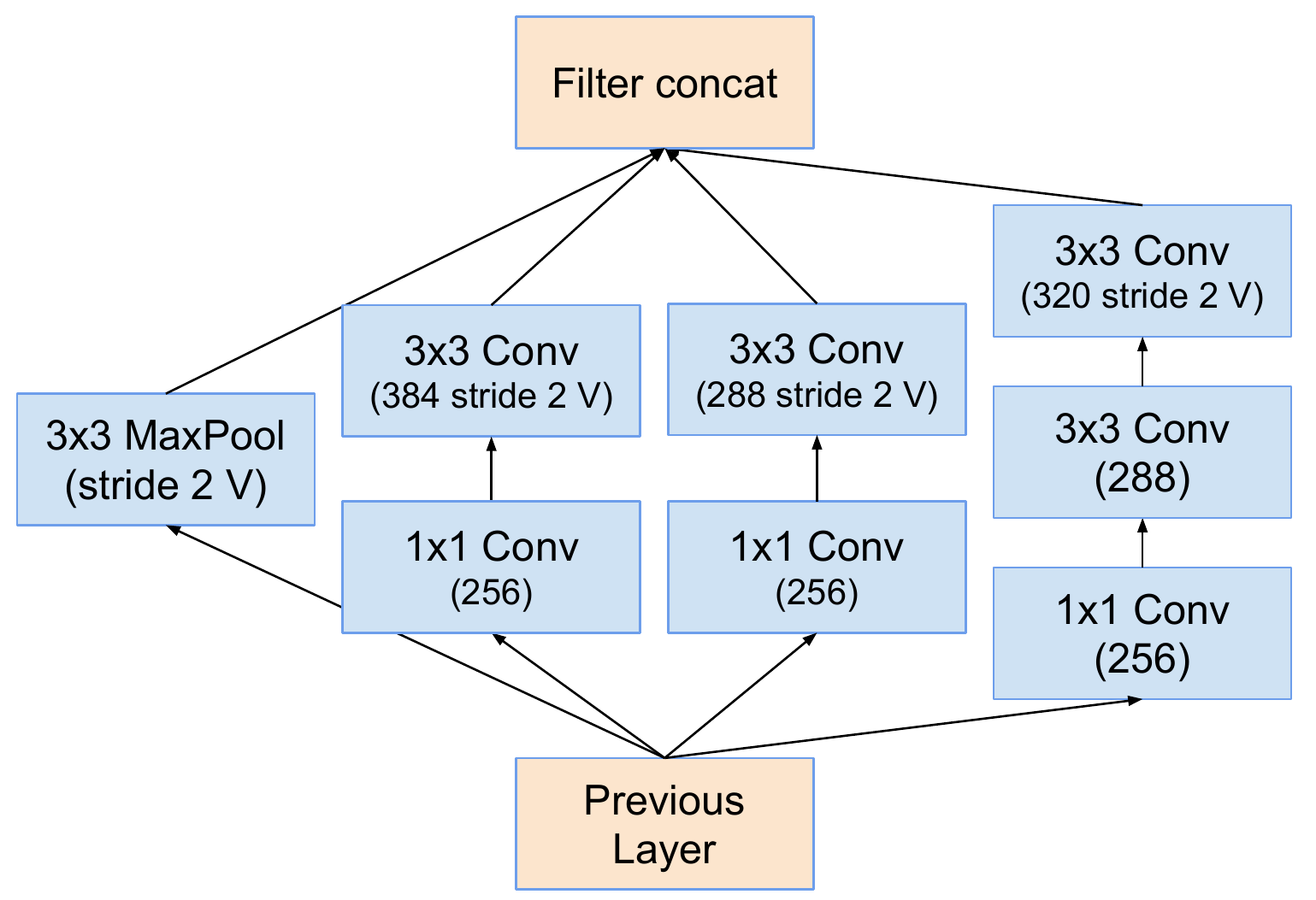}
\caption{The schema for $17\times 17$ to $8\times 8$ grid-reduction module.
  Reduction-B module used by the wider Inception-ResNet-v1 network
  in Figure~\ref{fig:resnetsmallschema}.
}
\label{fig:reductionto8resnetwide}
\end{figure}
\begin{figure}
\centering
\includegraphics[width=\linewidth]{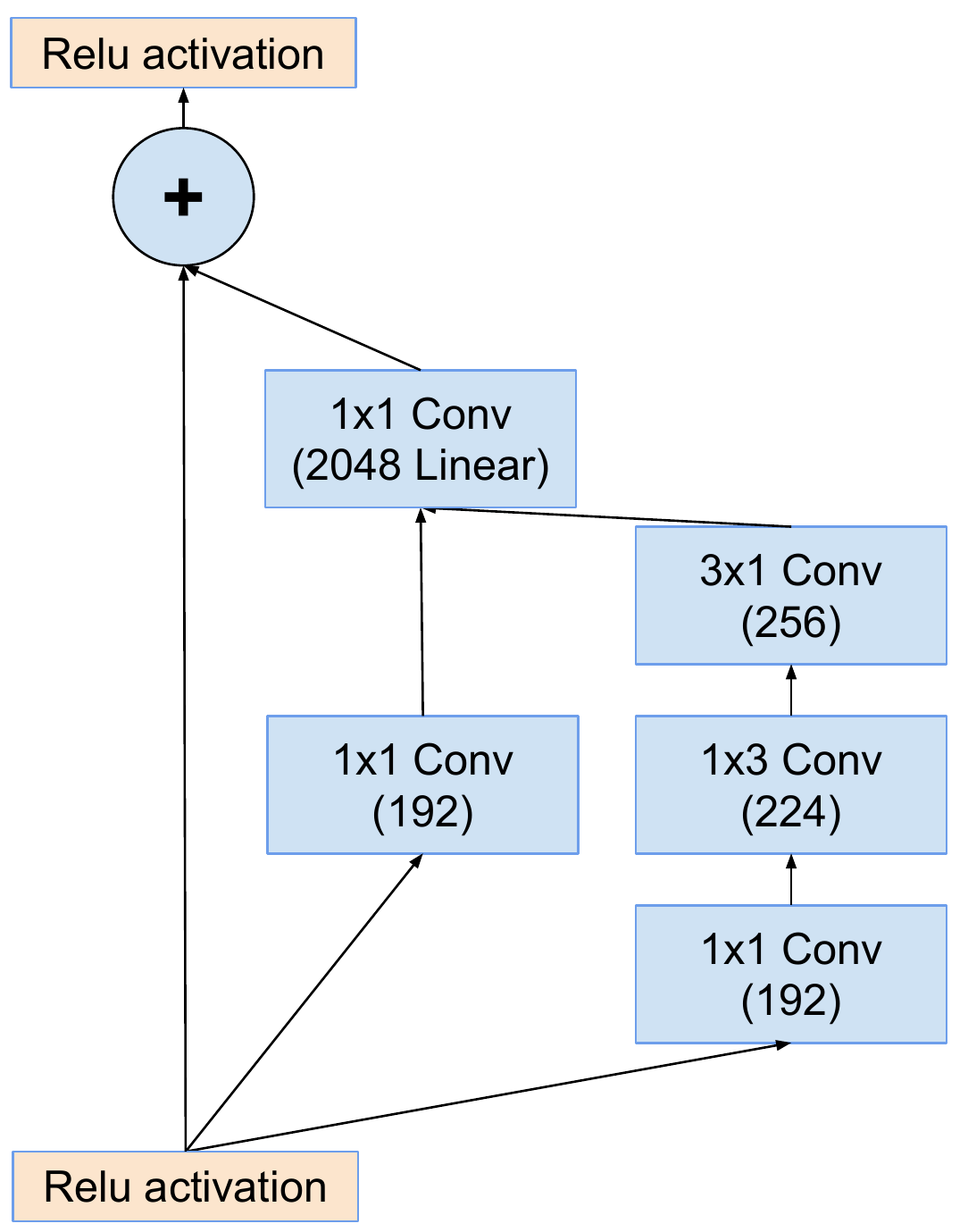}
\caption{The schema for $8\times 8$ grid  (Inception-ResNet-C) module of the
 Inception-ResNet-v2 network.}
\label{fig:resnetwide8x8module}
\end{figure}
\begin{table}
{\small
 \begin{center}
   \begin{tabular}[H]{|l|c|c|c|c|}
   \hline
   {\bf Network} & {$k$} & {$l$} & {$m$} & {$n$} \\
   \hline
   Inception-v4 & 192 & 224 & 256 & 384 \\
   Inception-ResNet-v1 & 192 & 192 & 256 & 384 \\
   Inception-ResNet-v2 & 256 & 256 & 384 & 384 \\
   \hline
   \end{tabular}
 \end{center}
 }
\caption{The number of filters of the Reduction-A module for the three
  Inception variants presented in this paper. The four numbers in the
  colums of the paper parametrize the four convolutions of Figure~\ref{fig:reductionto17} }
\label{reductionto17params}
\end{table}

\subsection{Scaling of the Residuals}
Also we found that if the number of filters exceeded 1000, the residual
variants started to exhibit instabilities and the network has just
``died'' early in the training, meaning that the last layer before the
average pooling started to produce only zeros after a few tens of thousands of
iterations. This could not be prevented, neither by lowering the learning
rate, nor by adding an extra batch-normalization to this layer.

\begin{figure}
\centering
\includegraphics[width=0.4\linewidth]{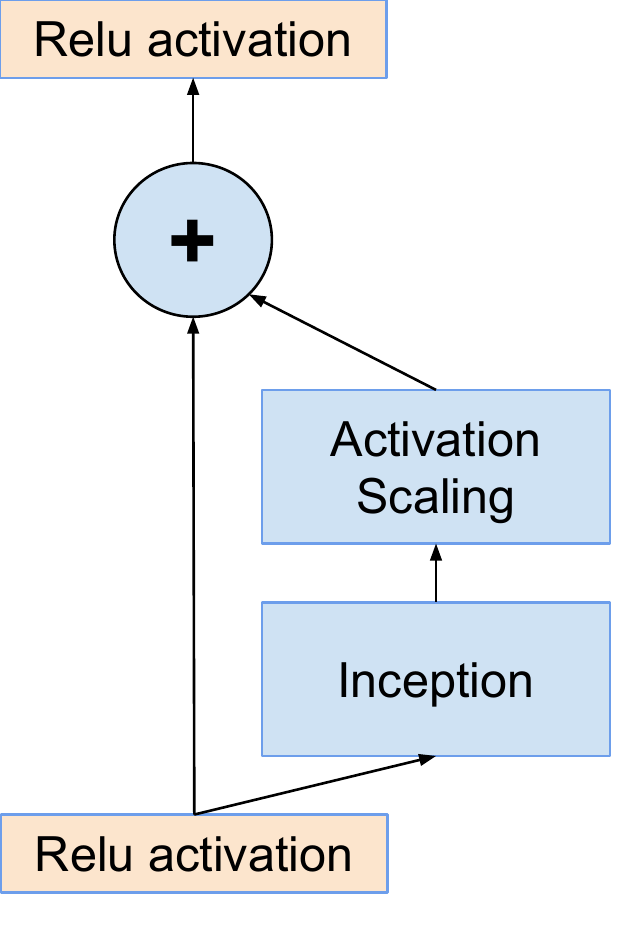}
\caption{The general schema for scaling combined Inception-resnet moduels.
  We expect that the same idea is useful in the general resnet case, where
instead of the Inception block an arbitrary subnetwork is used. The scaling
block just scales the last linear activations by a suitable constant, typically
around 0.1.}
\label{fig:resnetscaling}
\end{figure}

We found that scaling down the residuals before adding them to
the previous layer activation seemed to stabilize the training. In general
we picked some scaling factors between 0.1 and 0.3 to scale the residuals
before their being added to the accumulated layer activations
(cf. Figure~\ref{fig:resnetscaling}).

A similar instability was observed by He et al. in~\cite{he2015deep} in
the case of very deep residual networks and they suggested a two-phase
training where the first ``warm-up'' phase is done with very low learning
rate, followed by a second phase with high learning rata. We found that
if the number of filters is very high, then even a very low (0.00001) learning
rate is not sufficient to cope with the instabilities and the training with
high learning rate had a chance to destroy its effects. We found it much
more reliable to just scale the residuals.

Even where the scaling was not strictly necessary, it never
seemed to harm the final accuracy, but it helped to stabilize the training.

\section{Training Methodology}
We have trained our networks with stochastic gradient utilizing the
TensorFlow~\cite{tensorflow2015-whitepaper} distributed machine learning system
using $20$ replicas running each on a NVidia Kepler GPU.
Our earlier experiments used momentum~\cite{icml2013_sutskever13} with a
decay of $0.9$, while our best models were achieved using RMSProp~\cite{rmsprop}
with decay of $0.9$ and $\epsilon=1.0$. We used a learning rate of $0.045$,
decayed every two epochs using an exponential rate of $0.94$.
Model evaluations are performed using a running average of the parameters
computed over time.

\section{Experimental Results}

First we observe the top-1 and top-5 validation-error evolution of the
four variants during training. After the experiment was conducted, we have
found that our continuous evaluation was conducted on a subset of the
validation set which omitted about 1700 blacklisted entities due to poor
bounding boxes. It turned out that the omission should have been only performed
for the CLSLOC benchmark, but yields somewhat incomparable (more optimistic)
numbers when compared to other reports including some earlier reports by our
team. The difference is about 0.3\% for top-$1$ error and about 0.15\% for
the top-$5$ error. However, since the differences are consistent, we think
the comparison between the curves is a fair one.

On the other hand, we have rerun our multi-crop and ensemble results on the
complete validation set consisting of 50000 images. Also the final ensemble
result was also performed on the test set and sent to the ILSVRC test server
for validation to verify that our tuning did not result in an over-fitting.
We would like to stress that this final validation was done only once and we
have submitted our results only twice in the last year: once for the
BN-Inception paper and later during the ILSVR-2015 CLSLOC competition, so
we believe that the test set numbers constitute a true estimate of the
generalization capabilities of our model.

\begin{figure}
\centering
\includegraphics[width=\linewidth]{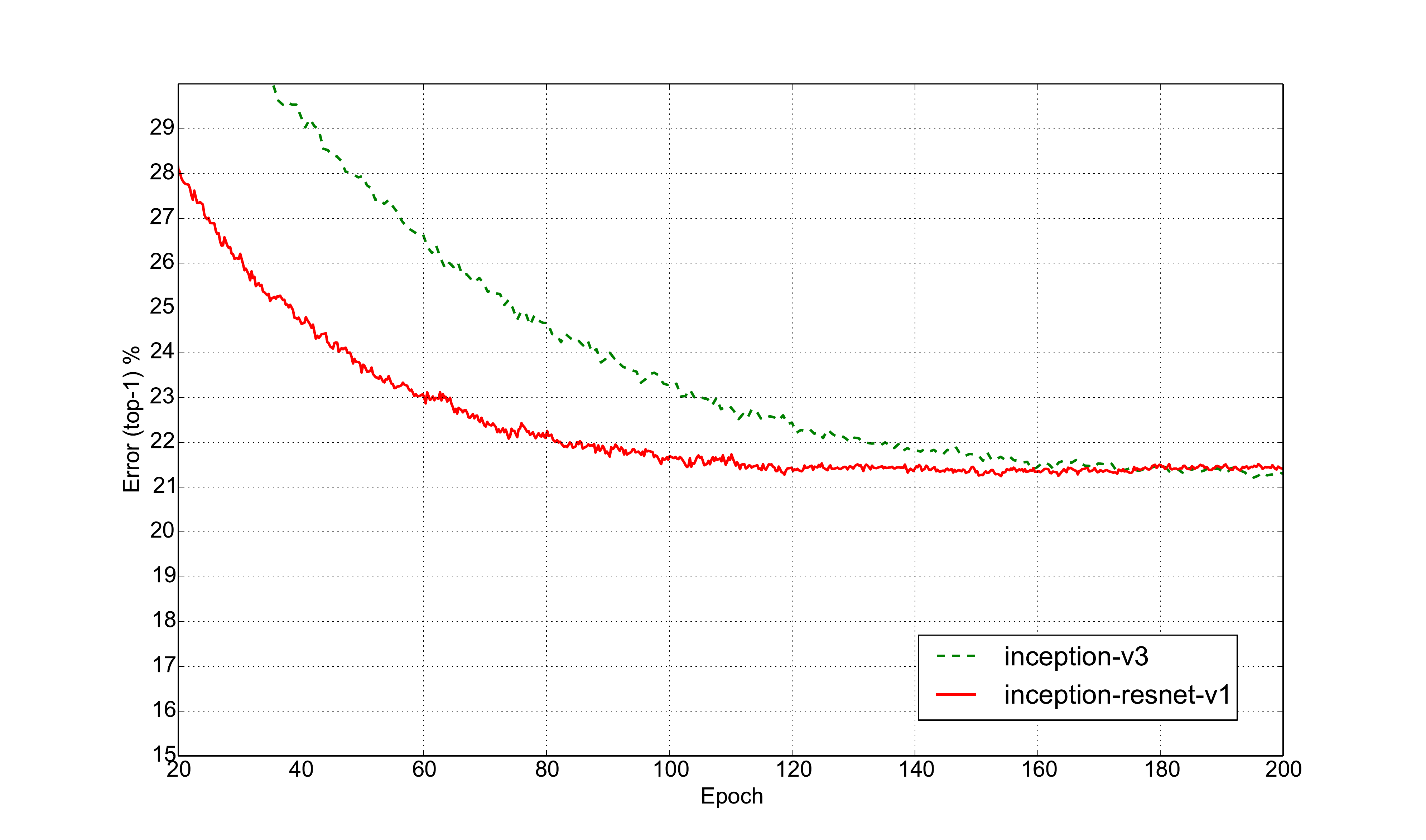}
\caption{Top-1 error evolution during training of pure Inception-v3 vs a
  residual network of similar computational cost. The evaluation is measured on
  a single crop on the non-blacklist images of the ILSVRC-2012 validation set.
  The residual model was training much faster, but reached
  slightly worse final accuracy than the traditional Inception-v3.
}
\label{fig:smalltop1}
\end{figure}

\begin{figure}
\centering
\includegraphics[width=\linewidth]{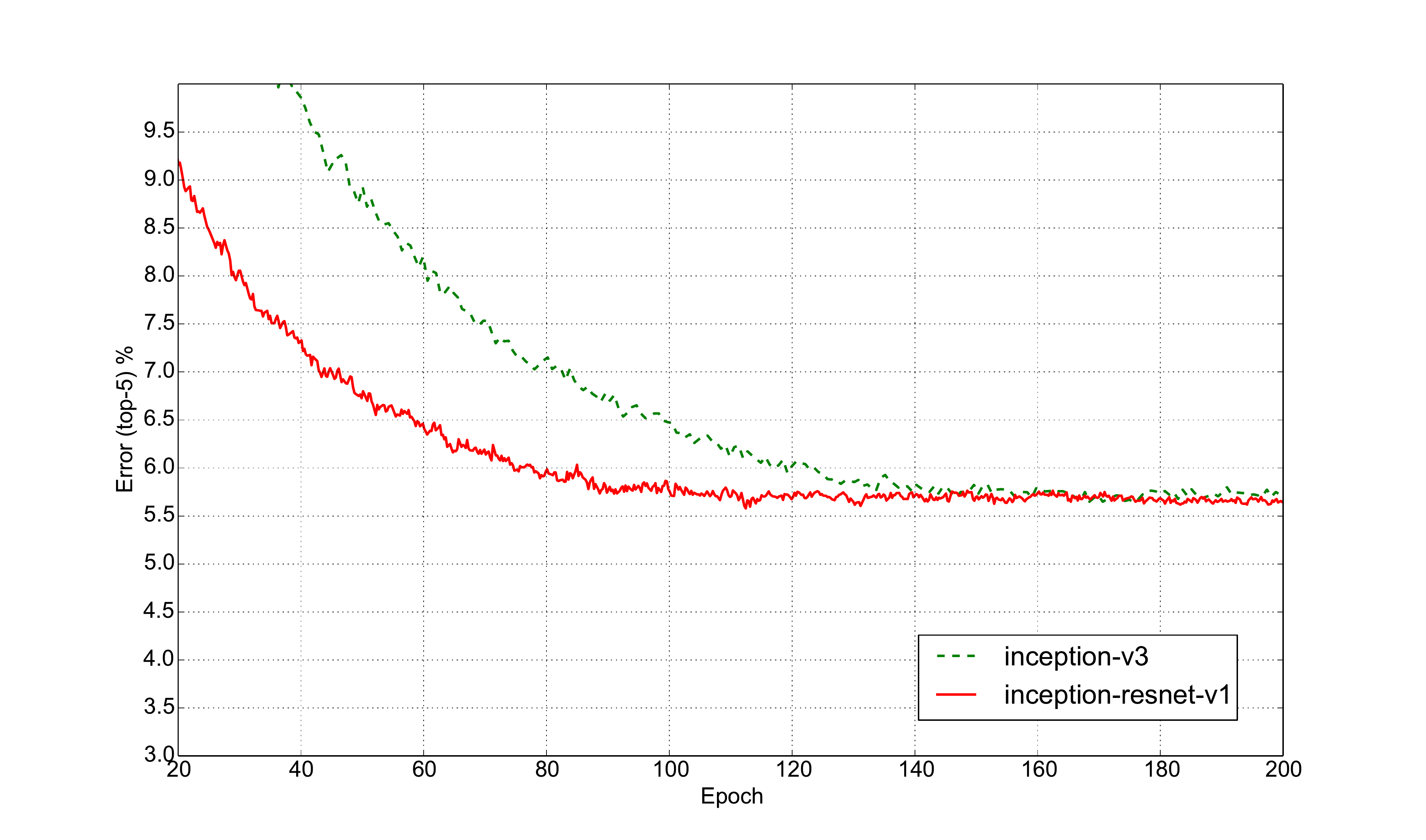}
\caption{Top-5 error evolution during training of pure Inception-v3 vs a
  residual Inception of similar computational cost. The evaluation is measured on
  a single crop on the non-blacklist images of the ILSVRC-2012 validation set.
  The residual version has trained much faster and reached slightly better final recall
  on the validation set.
}
\label{fig:smalltop5}
\end{figure}

\begin{figure}
\centering
\includegraphics[width=\linewidth]{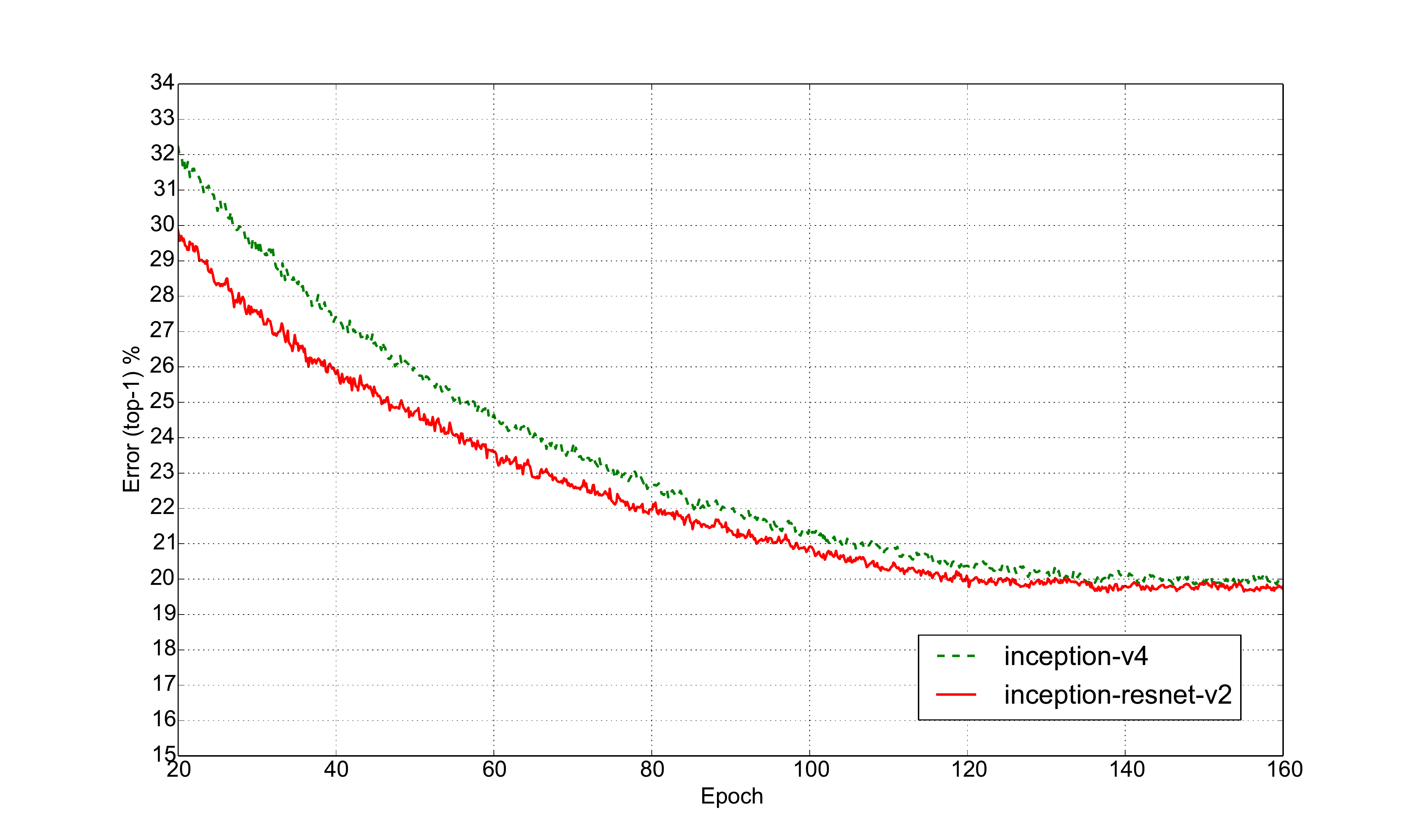}
\caption{Top-1 error evolution during training of pure Inception-v3 vs a
  residual Inception of similar computational cost. The evaluation is measured on
  a single crop on the non-blacklist images of the ILSVRC-2012 validation set.
  The residual version was training much faster and reached
  slightly better final accuracy than the traditional Inception-v4.
}
\label{fig:widetop1}
\end{figure}

\begin{figure}
\centering
\includegraphics[width=\linewidth]{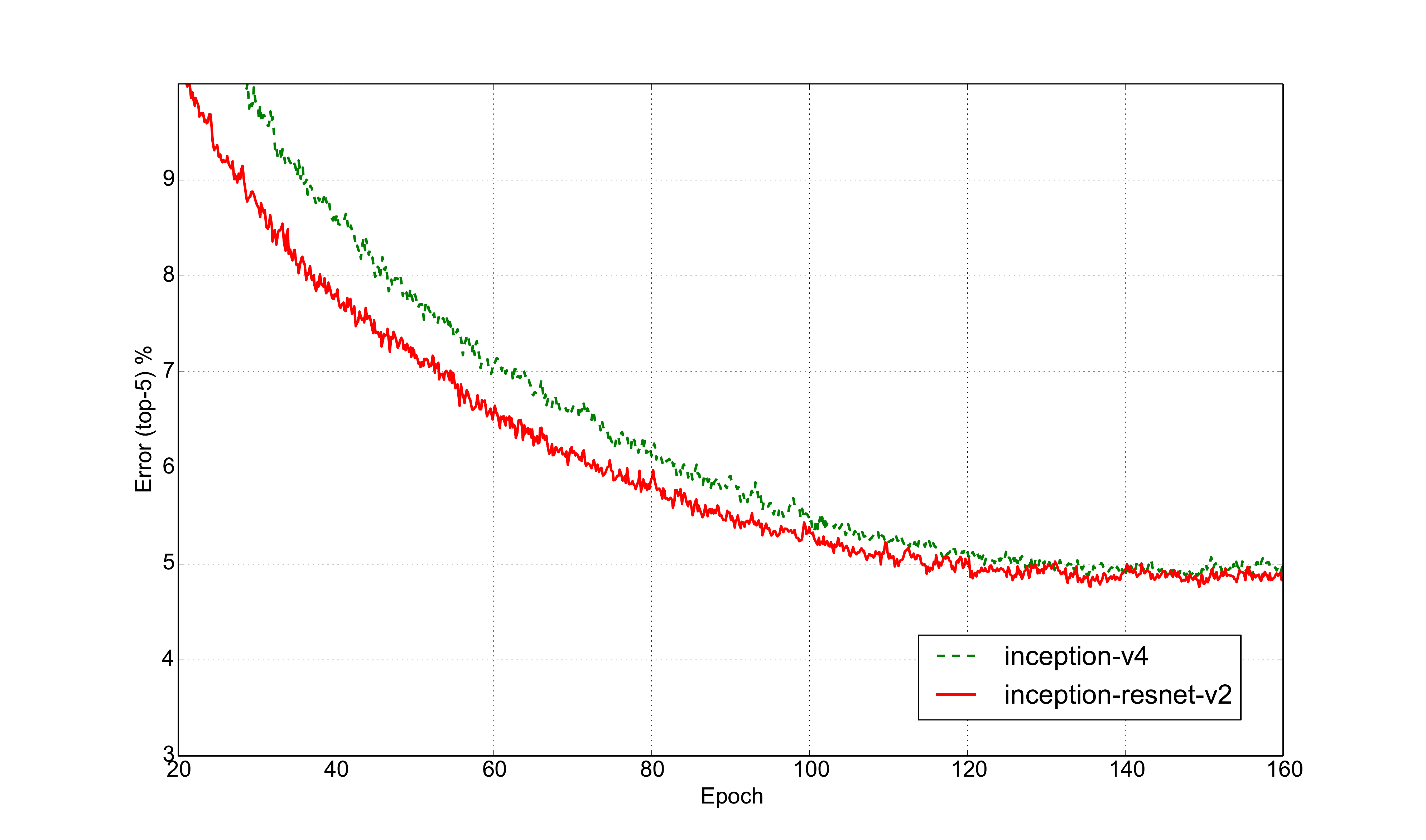}
\caption{Top-5 error evolution during training of pure Inception-v4 vs a
  residual Inception of similar computational cost. The evaluation is measured on
  a single crop on the non-blacklist images of the ILSVRC-2012 validation set.
  The residual version trained faster and reached slightly better final recall
  on the validation set.
}
\label{fig:widetop5}
\end{figure}

\begin{figure}
\centering
\includegraphics[width=\linewidth]{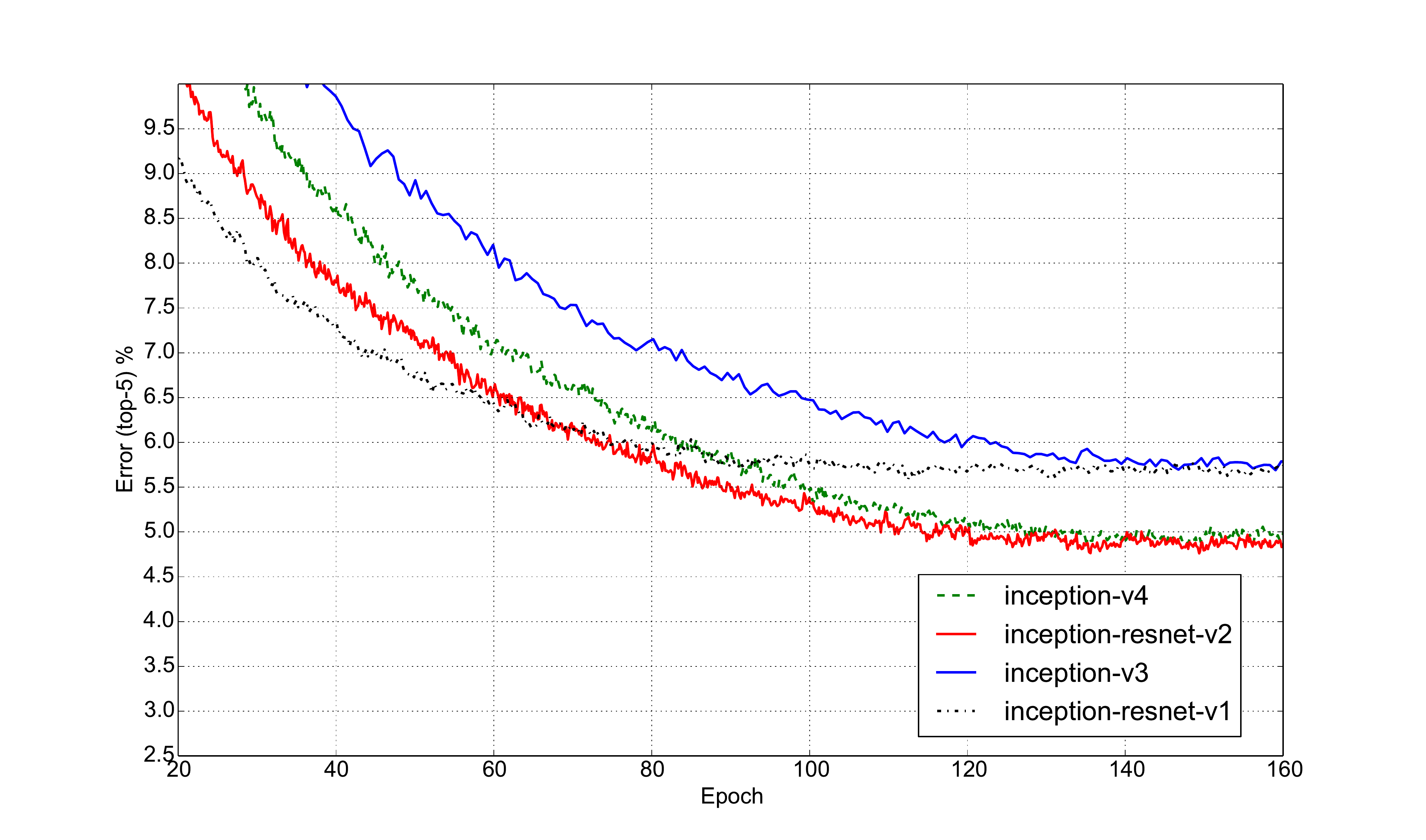}
\caption{Top-5 error evolution of all four models (single model, single crop).
  Showing the improvement due to larger model size. Although the residual
  version converges faster, the final accuracy seems to mainly depend on the
  model size.
}
\label{fig:alltop5}
\end{figure}

\begin{figure}
\centering
\includegraphics[width=\linewidth]{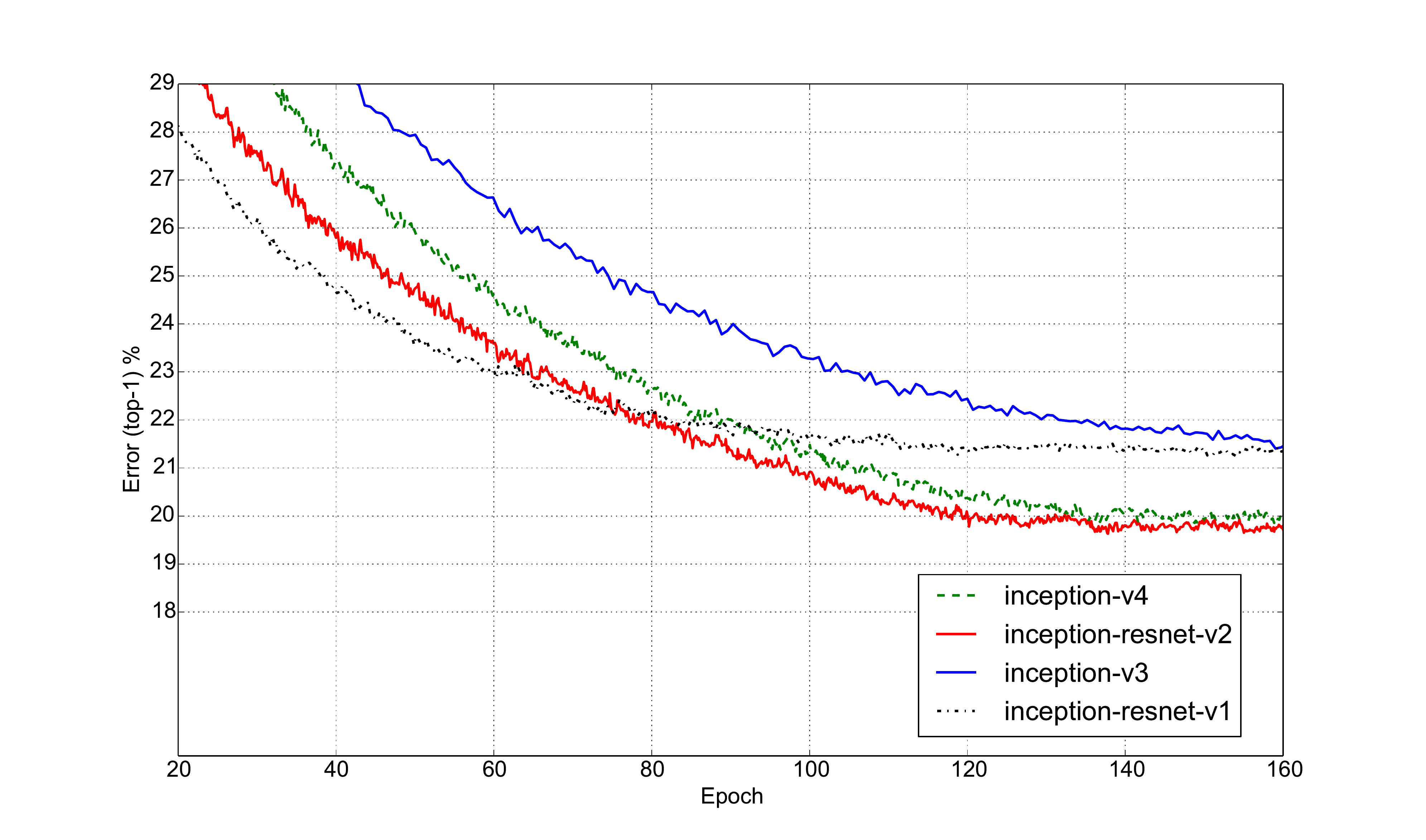}
\caption{Top-1 error evolution of all four models (single model, single crop).
  This paints a similar picture as the top-5 evaluation.
}
\label{fig:alltop1}
\end{figure}

Finally, we present some comparisons, between various versions of Inception
and Inception-ResNet. The models Inception-v3 and Inception-v4 are deep
convolutional networks not utilizing residual connections while
Inception-ResNet-v1 and Inception-ResNet-v2 are Inception style networks
that utilize residual connections instead of filter concatenation.

\begin{table}
{\small
 \begin{center}
   \begin{tabular}[H]{|l|c|c|}
   \hline
   {\bf Network} & {\bf Top-1 Error} & {\bf Top-5 Error} \\
   \hline
   BN-Inception~\cite{ioffe2015batch} & 25.2\% & 7.8\% \\
   Inception-v3~\cite{szegedy2015rethinking} & 21.2\% & 5.6\% \\
   Inception-ResNet-v1 & 21.3\% & 5.5\% \\
   Inception-v4 & 20.0\% & 5.0\% \\
   Inception-ResNet-v2 & 19.9\% & 4.9\% \\
   \hline
   \end{tabular}
 \end{center}
 }
\caption{Single crop - single model experimental results. Reported on the
non-blacklisted subset of the validation set of ILSVRC 2012.}
\label{singlesingle}
\end{table}
\begin{table}
{\small
 \begin{center}
   \begin{tabular}[H]{|l|c|c|c|}
   \hline
   {\bf Network} & {Crops} & {\bf Top-1 Error} & {\bf Top-5 Error} \\
   \hline
   ResNet-151~\cite{he2015deep} & 10 & 21.4\% & 5.7\% \\
   Inception-v3~\cite{szegedy2015rethinking} & 12 & 19.8\% & 4.6\% \\
   Inception-ResNet-v1 & 12 & 19.8\% & 4.6\% \\
   Inception-v4 & 12 & 18.7\% & 4.2\% \\
   Inception-ResNet-v2 & 12 & 18.7\% & 4.1\% \\
   \hline
   \end{tabular}
 \end{center}
 }
\caption{10/12 crops evaluations - single model experimental results.
  Reported on the
all 50000 images of the validation set of ILSVRC 2012.}
\label{multisingle}
\end{table}
\begin{table}
{\small
 \begin{center}
   \begin{tabular}[H]{|l|c|c|c|}
   \hline
   {\bf Network} & {Crops} & {\bf Top-1 Error} & {\bf Top-5 Error} \\
   \hline
   ResNet-151~\cite{he2015deep} & dense & 19.4\% & 4.5\% \\
   Inception-v3~\cite{szegedy2015rethinking} & 144 & 18.9\% & 4.3\% \\
   Inception-ResNet-v1 & 144 & 18.8\% & 4.3\% \\
   Inception-v4 & 144 & 17.7\% & 3.8\% \\
   Inception-ResNet-v2 & 144 & 17.8\% & 3.7\% \\
   \hline
   \end{tabular}
 \end{center}
 }
\caption{144 crops evaluations - single model experimental results.
  Reported on the all 50000 images of the validation set of ILSVRC 2012.}
\label{manysingle}
\end{table}
\begin{table}
{\small
 \begin{center}
   \begin{tabular}[H]{|l|c|c|c|}
   \hline
   {\bf Network} & {Models} & {\bf Top-1 Error} & {\bf Top-5 Error} \\
   \hline
   ResNet-151~\cite{he2015deep} & 6 & -- & 3.6\% \\
   Inception-v3~\cite{szegedy2015rethinking} & 4 & 17.3\% & 3.6\% \\
   \stackanchor{Inception-v4 + }{$3\times$ Inception-ResNet-v2} & 4 &
   16.5\% & 3.1\% \\
   \hline
   \end{tabular}
 \end{center}
 }
\caption{Ensemble results with 144 crops/dense evaluation.
  Reported on the all 50000 images of the validation set of ILSVRC 2012.
  For Inception-v4(+Residual), the ensemble consists of one pure Inception-v4
  and three Inception-ResNet-v2 models and were evaluated both on the
  validation and on the test-set. The test-set performance was
  $3.08\%$ top-5 error verifying that we don't over-fit on the validation
  set.
}
\label{manyensemble}
\end{table}

Table~\ref{singlesingle} shows the single-model, single crop top-1 and
top-5 error of the various architectures on the validation set.

Table~\ref{multisingle} shows the performance of the various models with a
small number of crops: 10 crops for ResNet as was reported
in~\cite{he2015deep}), for the Inception variants, we have used the 12 crops
evaluation as as described in~\cite{szegedy2015going}.

Table~\ref{manysingle} shows the single model performance of the various
models using. For residual network the dense evaluation result is reported
from~\cite{he2015deep}. For the inception networks, the 144 crops strategy
was used as described in~\cite{szegedy2015going}.

Table~\ref{manyensemble} compares ensemble results. For the pure residual
network the 6 models dense evaluation result is reported
from~\cite{he2015deep}. For the inception networks 4 models were ensembled
using the 144 crops strategy as described in~\cite{szegedy2015going}.

\section{Conclusions}

We have presented three new network architectures in detail:

\begin{itemize}
  \item Inception-ResNet-v1: a hybrid Inception version that has a
    similar computational cost to Inception-v3
    from~\cite{szegedy2015rethinking}.
  \item Inception-ResNet-v2: a costlier hybrid Inception version with
    significantly improved recognition performance.
  \item Inception-v4: a pure Inception variant without residual connections
    with roughly the same recognition performance as Inception-ResNet-v2.
\end{itemize}

We studied how the introduction of residual connections leads to dramatically
improved training speed for the Inception architecture. Also our latest models
(with and without residual connections) outperform all our previous networks,
just by virtue of the increased model size.



{\small
\bibliographystyle{ieee}
\bibliography{references}

\begin{thebibliography}{10}\itemsep=-1pt

\bibitem{tensorflow2015-whitepaper}
M.~Abadi, A.~Agarwal, P.~Barham, E.~Brevdo, Z.~Chen, C.~Citro, G.~S. Corrado,
  A.~Davis, J.~Dean, M.~Devin, S.~Ghemawat, I.~Goodfellow, A.~Harp, G.~Irving,
  M.~Isard, Y.~Jia, R.~Jozefowicz, L.~Kaiser, M.~Kudlur, J.~Levenberg,
  D.~Man\'{e}, R.~Monga, S.~Moore, D.~Murray, C.~Olah, M.~Schuster, J.~Shlens,
  B.~Steiner, I.~Sutskever, K.~Talwar, P.~Tucker, V.~Vanhoucke, V.~Vasudevan,
  F.~Vi\'{e}gas, O.~Vinyals, P.~Warden, M.~Wattenberg, M.~Wicke, Y.~Yu, and
  X.~Zheng.
\newblock {TensorFlow}: Large-scale machine learning on heterogeneous systems,
  2015.
\newblock Software available from tensorflow.org.

\bibitem{dean2012large}
J.~Dean, G.~Corrado, R.~Monga, K.~Chen, M.~Devin, M.~Mao, A.~Senior, P.~Tucker,
  K.~Yang, Q.~V. Le, et~al.
\newblock Large scale distributed deep networks.
\newblock In {\em Advances in Neural Information Processing Systems}, pages
  1223--1231, 2012.

\bibitem{dong2014learning}
C.~Dong, C.~C. Loy, K.~He, and X.~Tang.
\newblock Learning a deep convolutional network for image super-resolution.
\newblock In {\em Computer Vision--ECCV 2014}, pages 184--199. Springer, 2014.

\bibitem{girshick2014rcnn}
R.~Girshick, J.~Donahue, T.~Darrell, and J.~Malik.
\newblock Rich feature hierarchies for accurate object detection and semantic
  segmentation.
\newblock In {\em Proceedings of the IEEE Conference on Computer Vision and
  Pattern Recognition ({CVPR})}, 2014.

\bibitem{he2015deep}
K.~He, X.~Zhang, S.~Ren, and J.~Sun.
\newblock Deep residual learning for image recognition.
\newblock {\em arXiv preprint arXiv:1512.03385}, 2015.

\bibitem{ioffe2015batch}
S.~Ioffe and C.~Szegedy.
\newblock Batch normalization: Accelerating deep network training by reducing
  internal covariate shift.
\newblock In {\em Proceedings of The 32nd International Conference on Machine
  Learning}, pages 448--456, 2015.

\bibitem{karpathy2014large}
A.~Karpathy, G.~Toderici, S.~Shetty, T.~Leung, R.~Sukthankar, and L.~Fei-Fei.
\newblock Large-scale video classification with convolutional neural networks.
\newblock In {\em Computer Vision and Pattern Recognition (CVPR), 2014 IEEE
  Conference on}, pages 1725--1732. IEEE, 2014.

\bibitem{krizhevsky2012imagenet}
A.~Krizhevsky, I.~Sutskever, and G.~E. Hinton.
\newblock Imagenet classification with deep convolutional neural networks.
\newblock In {\em Advances in neural information processing systems}, pages
  1097--1105, 2012.

\bibitem{lin2013network}
M.~Lin, Q.~Chen, and S.~Yan.
\newblock Network in network.
\newblock {\em arXiv preprint arXiv:1312.4400}, 2013.

\bibitem{long2015fully}
J.~Long, E.~Shelhamer, and T.~Darrell.
\newblock Fully convolutional networks for semantic segmentation.
\newblock In {\em Proceedings of the IEEE Conference on Computer Vision and
  Pattern Recognition}, pages 3431--3440, 2015.

\bibitem{russakovsky2014imagenet}
O.~Russakovsky, J.~Deng, H.~Su, J.~Krause, S.~Satheesh, S.~Ma, Z.~Huang,
  A.~Karpathy, A.~Khosla, M.~Bernstein, et~al.
\newblock Imagenet large scale visual recognition challenge.
\newblock 2014.

\bibitem{simonyan2014very}
K.~Simonyan and A.~Zisserman.
\newblock Very deep convolutional networks for large-scale image recognition.
\newblock {\em arXiv preprint arXiv:1409.1556}, 2014.

\bibitem{icml2013_sutskever13}
I.~Sutskever, J.~Martens, G.~Dahl, and G.~Hinton.
\newblock On the importance of initialization and momentum in deep learning.
\newblock In {\em Proceedings of the 30th International Conference on Machine
  Learning (ICML-13)}, volume~28, pages 1139--1147. JMLR Workshop and
  Conference Proceedings, May 2013.

\bibitem{szegedy2015going}
C.~Szegedy, W.~Liu, Y.~Jia, P.~Sermanet, S.~Reed, D.~Anguelov, D.~Erhan,
  V.~Vanhoucke, and A.~Rabinovich.
\newblock Going deeper with convolutions.
\newblock In {\em Proceedings of the IEEE Conference on Computer Vision and
  Pattern Recognition}, pages 1--9, 2015.

\bibitem{szegedy2015rethinking}
C.~Szegedy, V.~Vanhoucke, S.~Ioffe, J.~Shlens, and Z.~Wojna.
\newblock Rethinking the inception architecture for computer vision.
\newblock {\em arXiv preprint arXiv:1512.00567}, 2015.

\bibitem{rmsprop}
T.~Tieleman and G.~Hinton.
\newblock Divide the gradient by a running average of its recent magnitude.
\newblock COURSERA: Neural Networks for Machine Learning, 4, 2012.
\newblock Accessed: 2015-11-05.

\bibitem{toshev2014deeppose}
A.~Toshev and C.~Szegedy.
\newblock Deeppose: Human pose estimation via deep neural networks.
\newblock In {\em Computer Vision and Pattern Recognition (CVPR), 2014 IEEE
  Conference on}, pages 1653--1660. IEEE, 2014.

\bibitem{wang2013learning}
N.~Wang and D.-Y. Yeung.
\newblock Learning a deep compact image representation for visual tracking.
\newblock In {\em Advances in Neural Information Processing Systems}, pages
  809--817, 2013.

\end{thebibliography}
}

\end{document}